\documentclass{article}
\usepackage[preprint, nonatbib]{neurips_2024}

\usepackage[utf8]{inputenc}
\usepackage[T1]{fontenc}
\usepackage{hyperref}
\usepackage{url}
\usepackage{booktabs}
\usepackage{amsfonts}
\usepackage{nicefrac}
\usepackage{microtype}
\usepackage{xcolor}

\usepackage{import}
\usepackage{dirtytalk}
\usepackage{subcaption}
\usepackage{adjustbox}
\usepackage{subcaption}
\usepackage{enumitem}
\usepackage{amsthm}

\usepackage{amssymb}
\usepackage{mathtools}

\usepackage{colortbl}
\usepackage{xcolor}
\usepackage{tcolorbox}
\usepackage{color,soul}
\usepackage{siunitx}

\usepackage{booktabs} 
\usepackage{float}
\usepackage{makecell}
\usepackage{subcaption}
\usepackage{multirow}

\definecolor{firstplace}{RGB}{230, 240, 255}
\definecolor{secondplace}{RGB}{240, 255, 240}
\definecolor{thirdplace}{RGB}{255, 240, 230}
\newcommand{\coloredcell}[2]{
  \begingroup
  \setlength{\fboxsep}{0pt}
  \colorbox{#1}{
    \parbox[c][1.3em][c]{7em}{\centering #2}
  }
  \endgroup
}

\usepackage{mdframed}
\theoremstyle{plain}
\newtheorem{theorem}{Theorem}[section]
\theoremstyle{definition}
\newtheorem{definition}[theorem]{Definition}

\usepackage[ruled, linesnumbered]{algorithm2e}
\usepackage{wrapfig}

\usepackage[numbers]{natbib}

\usepackage{cleveref}

\makeatletter
\newcommand{\myfnsymbol}[1]{
  \expandafter\@myfnsymbol\csname c@#1\endcsname
}
\newcommand{\@myfnsymbol}[1]{
  \ifcase #1
  \or 1
  \or 2
  \or \TextOrMath{\textasteriskcentered}{*}
  \or \TextOrMath{\textasteriskcentered}{*}\TextOrMath{\textasteriskcentered}{*}
  \or \TextOrMath{\textdagger}{\dagger}
  \or \TextOrMath{\textasteriskcentered}{*},\TextOrMath{\textasteriskcentered}{*}\TextOrMath{\textasteriskcentered}{*}
  \fi
}
\newcommand{\affiliationA}{\@myfnsymbol{1}}
\newcommand{\affiliationB}{\@myfnsymbol{2}}
\newcommand{\correspondingA}{\@myfnsymbol{3}}
\newcommand{\biequalcontributor}{\@myfnsymbol{4}}
\newcommand{\equalcontributor}{\@myfnsymbol{5}}

\title{Sheaf HyperNetworks for Personalized \\ Federated Learning}

\author{
\large Bao Nguyen\textsuperscript{\correspondingA,\affiliationA}
\And
\large Lorenzo Sani\textsuperscript{\affiliationA,\affiliationB}
\And
\large Xinchi Qiu\textsuperscript{\affiliationA}
\And
\large Pietro Liò\textsuperscript{\affiliationA}~~~~~
\large Nicholas D.\ Lane\textsuperscript{\affiliationA,\affiliationB}
\\\\
\textsuperscript{\affiliationA}{Department of Computer Science and Technology, University of Cambridge}
\\
\textsuperscript{\affiliationB}{Flower Labs}
\\
\textsuperscript{\correspondingA}{Corresponding author: \texttt{btn21@cam.ac.uk}}
}

\begin{document}

\maketitle

\begin{abstract}
Graph hypernetworks (GHNs), constructed by combining graph neural networks (GNNs) with hypernetworks (HNs), leverage relational data across various domains such as neural architecture search, molecular property prediction and federated learning. Despite GNNs and HNs being individually successful, we show that GHNs present problems compromising their performance, such as over-smoothing and heterophily. Moreover, we cannot apply GHNs directly to personalized federated learning (PFL) scenarios, where a priori client relation graph may be absent, private, or inaccessible. To mitigate these limitations in the context of PFL, we propose a novel class of HNs, sheaf hypernetworks (SHNs), which combine cellular sheaf theory with HNs to improve parameter sharing for PFL. We thoroughly evaluate SHNs across diverse PFL tasks, including multi-class classification, traffic and weather forecasting. Additionally, we provide a methodology for constructing client relation graphs in scenarios where such graphs are unavailable. We show that SHNs consistently outperform existing PFL solutions in complex non-IID scenarios. While the baselines' performance fluctuates depending on the task, SHNs show improvements of up to 2.7\% in accuracy and 5.3\% in lower mean squared error over the best-performing baseline.
\end{abstract}

\section{Introduction}

Hypernetworks (HNs) \cite{ha_hypernetworks_2016} are a class of neural networks that generate parameters for a target network. They have emerged as a practical solution across tasks ranging from neural architecture search \cite{brock2017smash, zhang_graph_2020}, molecular property prediction \cite{nachmani2020molecule, wu2024pacia} to federated learning \cite{shamsian_personalized_2021, yi2023pfedkt, litany2022federated}.
\citet{zhang_graph_2020} proposed to couple HNs with graph neural networks (GNNs) \cite{zhang_graph_2020} operated by graph convolutions \cite{kipf_semi-supervised_2017}.
Such a new class of HNs is called graph hypernetworks (GHN) and can leverage underlying graph relationships in the input space.
Although GHNs are effective, they may present GNN-related issues such as (a) over-smoothing due to increasing message-passing layers, which leads to node embeddings converging to the exact representation \cite{cai_note_2020, rusch2023survey}, and (b) heterophily, whereby nodes of different classes are linked \cite{barbero_sheaf_attention_2022,rusch2023survey}.
These problems result in a significant signal loss, which ultimately causes performance issues for the task objective.
Thus, relying on Graph Convolutional Networks (GCNs) \cite{kipf_semi-supervised_2017} for message passing in GHNs limits the depth of interaction~\cite{rusch2023survey}.
This limitation highlights the critical need for alternative solutions supporting deeper, more effective message-passing mechanisms without compromising model performance.

Federated learning (FL) \citep{mcmahan_communication-efficient_2023, konecny_federated_2015} is a privacy-preserving distributed learning paradigm that aims to learn a global model using private distributed data collaboratively.
The data is privately held by the clients, i.e.,~the entities collecting it, and never shared in the federation.
FL instead exchanges and aggregates model updates partially trained at clients iteratively.
Data heterogeneity among clients represents one of the main challenges of FL \citep{long_fl_survey, Li_2023}, degrading the global model's performance. Personalized federated learning (PFL) \citep{Tan_2023} offers an alternative training paradigm that delivers unique parameters to the clients in the federation based on their local data distribution while incorporating effective parameter sharing across clients.
However, data heterogeneity negatively impacts PFL's performance as it is often difficult to intercept the underlying relationship between clients whose private data allows for intelligent parameter sharing \cite{lee2024fedl2p}.

\citet{hansen_sheaf_2020, bodnar_neural_2023} equipped GNNs with cellular sheaves.
Such abstract algebraic structures enrich the expressivity of graph representation learning and extend the GNN literature toward modeling more complex interactions. Cellular sheaves can provide graphs with more powerful and flexible mathematical representations by associating vector spaces to nodes and enabling a new diffusion process through the sheaf Laplacian.
The natural extension of the GNNs to sheaves, i.e., sheaf neural networks (SNNs) \cite{hansen_sheaf_2020}, demonstrate the ability to mitigate the problems of over-smoothing and heterophily \cite{bodnar_neural_2023, barbero_sheaf_attention_2022, duta_sheaf_2023}.
Therefore, in this work, we introduce a novel class of HNs called Sheaf HyperNetworks (SHNs) to address the limitations mentioned above in GHNs, which combines the advantages of cellular sheaf theory \cite{hansen_toward_2019} with HNs. 

Our contributions are threefold:

\begin{enumerate}

    \item We introduce \textbf{Sheaf HyperNetworks (SHNs)} to mitigate GHNs' limitations, such as over-smoothing, heterophily, and lack of expressivity. We evaluate the effectiveness of SHNs on the task of personalized federated learning (PFL), where parameter sharing is challenging, and learning solid relationships between clients requires overcoming the expressivity limitations of GNNs. We show that SHN is effective under many PFL tasks, such as multi-class classification, traffic, and weather forecasting. Our model outperforms previous solutions across multiple heterogenous benchmarks, achieving 2.7\% higher accuracy and 5.3\% lower mean squared error over the best-performing baseline.
    
    \item We propose a \textbf{novel methodology} to construct a \textbf{client relation graph} using the learned client embeddings from a hypernetwork. Such a graph enables effective parameter sharing across comparable clients. It allows graph-based hypernetwork methods, such as SHN, to be applied to settings where relational information between clients is not available.

    \item We provide an extensive evaluation to measure the \textbf{robustness} of SHN compared to GHN. In our assessment, we measure the performance of PFL, varying three critical parameters: the number of message-passing layers, the number of nearest neighbors, and the cosine threshold used to construct our client relation graph.

\end{enumerate}

\section{Background}

In this section, we describe the background of our work.
First, we discuss federated learning (FL) and its variant, personalized federated learning (PFL).
Subsequently, we detail the motivation and utility of hypernetworks.
Finally, we discuss the components of a cellular sheaf, their expressivity, and their ability to mitigate over-smoothing and heterophily.

\textbf{HyperNetworks.}
Training large-scale neural networks via traditional gradient descent methods is notably arduous owing to the intricate task of navigating expansive parameter spaces. In response, \citet{ha_hypernetworks_2016} introduced hypernetworks (HNs), which generate parameters for a target network (TN) obtaining more efficient training by reducing the parameter search space \cite{chauhan_brief_2023}. HNs have shown competitive performance on tasks with variants conditioned on specific input spaces $H$, generalizing to any neural network architecture and learning objective. Typically composed of multi-layer perceptrons (MLPs) for each TN layer, HNs act as coordinate mapping functions, projecting vectors from the smaller embedding space 
$H$ onto the TN's expansive parameter space.

\begin{definition}
\textup{We can define the optimization strategy for a hypernetwork (HN) and its coupled target network (TN) as:}
\begin{equation}
\text {HN:} \quad \min _{\psi_\mathcal{H}} \mathcal{T}(X ; \Theta)=\mathcal{T}(X ; \mathcal{HN}(X ; \psi_\mathcal{H})), \quad \text {TN:} \quad \min _{\Theta} \mathcal{T}(D ; \Theta) \text {. }
\end{equation}
\textup{A hypernetwork $\mathcal{HN}$ is equipped with parameters $\psi_{\mathcal{H}}$ and conditioned by the input space $X$. It aims to produce parameters $\theta \in \Theta$ for the target network $\mathcal{T}$. As shown, $\mathcal{T}$ remains unchanged: parameterized by $\Theta$ and conditioned by the input $D$.}
\end{definition}
\vspace{-0.5em}

The composition of a graph neural network (GNN) and an HN induces a relational inductive bias and enables the model to discern and leverage relational information in the graph.
\citet{zhang_graph_2020} proposed graph hypernetworks (GHNs) for neural architecture search by operating on a model's computation graph to generate parameters.
To the best of our knowledge, GHNs' implementations have always been prone to over-smoothing \cite{rusch2023survey} and heterophily because of their dependency on graph convolutional message-passing layers \cite{kipf_semi-supervised_2017}. We show that these phenomena, presenting a significant challenge for GNNs, also compromise GHN's effectiveness (\Cref{sec:effectiveness_of_sheaves}).
We address these challenges by introducing an adaptation in which the GNN is equipped with cellular sheaves, leading to our model, Sheaf HyperNetowork (SHN), for PFL.

\textbf{Cellular Sheaves.}
Cellular sheaves in GNNs \cite{hansen_sheaf_2020, bodnar_neural_2023, barbero_sheaf_attention_2022} are mathematical objects that equip a graph with spaces over its nodes and edges, called \emph{stalks}, and linear \emph{restriction maps} between these spaces. Recent works have explored the composition of cellular sheaves in the context of attention \cite{barbero_sheaf_attention_2022}, positional encodings \cite{he_sheaf-based_2023} and hypergraphs \cite{duta_sheaf_2023}, with few applications \cite{purificato_sheaf4rec_2024} beyond graph benchmark datasets. This section describes the building blocks and components needed to perform message-passing on a GNN equipped with a cellular sheaf.

\begin{definition}
\textup{A cellular sheaf $\mathcal{F}$ on an un-directed graph $G = (V, E)$ is defined as a triple $\left\langle\mathcal{F}(v), \mathcal{F}(e), \mathcal{F}_{v \unlhd e}\right\rangle$, where:}
\begin{enumerate}[nosep]
    \item \textup{$\mathcal{F}(v)$ are \emph{vertex stalks}: vector spaces associated with each node $v$},
    \item \textup{$\mathcal{F}(e)$ are \emph{edge stalks}: vector spaces associated with each edge $e$,}
    \item \textup{$\mathcal{F}_{v \unlhd e}$: $\mathcal{F}(v) \rightarrow \mathcal{F}(e)$ are linear \emph{restriction maps}: for each incident node-edge pair $v \unlhd e$.}
\end{enumerate}
\label{def:cellular_sheaf}
\end{definition}

\citet{hansen_opinion_2020} describe this from the perspective of opinion dynamics, whereby the node's \emph{stalks} are thought of as the private space of opinions and the edge's \emph{stalks} as a public discourse space. The linear \emph{restriction maps} $\mathcal{F}_{v \unlhd e}$ facilitate the flow of information between these spaces. We define the space of node and edge \emph{stalks}, as the space of $0$-cochains $C^0(G, \mathcal{F})$ and $1$-cochains $C^1(G, \mathcal{F})$, respectively.
Subsequently, a linear co-boundary map $\delta: C^0(G, \mathcal{F}) \rightarrow C^1(G, \mathcal{F})$ can be constructed to map the space of $0$-cochains and $1$-cochains together.
Such a map can be seen as the measure of disagreement between these \emph{spaces}. Through $\delta$, we can construct the symmetrically normalized \emph{sheaf Laplacian} $\Delta_{\mathcal{F}}$, which allows for sheaf diffusion over the cellular sheaf-equipped graph. In \Cref{appendix:cellular_sheaf_diffusion}, we detail the definition of each component described here.

\textbf{Mitigating Oversmoothing.}
For problem definitions in which maintaining the heterogeneity of the node representation is crucial, GNNs may suffer from the problem of over-smoothing.
Contrary to standard neural network models, GNNs don't gain expressivity by increasing the number of layers; instead, node embeddings converge to the same representation.
In other words, when the information in the immediate neighborhood is insufficient, nodes need to communicate with others that are $k$-hops apart, so $k$ GNN layers are needed.
However, \citet{cai_note_2020} showed that increasing $k$ exacerbates over-smoothing.

Sheaves have been shown to mitigate the effects of over-smoothing, subsequently improving task performance.
We extensively discuss this phenomenon and its relationship with sheaves in \Cref{appendix:oversmoothing}.
Several other models have been specifically designed to address over-smoothing.
For instance, GCNII \cite{chen2020simple} used initial residual connections combined with identity mappings to maintain node distinctiveness, and PairNorm \cite{zhao2020pairnorm} applied a normalization method to node features.
However, GNNs equipped with cellular sheaves are competitive with, or even outperform, both methods in mitigating over-smoothing and overall performance \cite{bodnar_neural_2023, barbero_sheaf_attention_2022, barbero_sheaf_2022}.

\textbf{Federated Learning.}
The goal of federated optimization \cite{mcmahan_communication-efficient_2023, konecny_federated_2015} is to minimize the empirical risk across all clients when equipped with a set of \emph{global model} parameters $\theta$: $\min_{\theta} l(\theta) = \sum_{m=1}^{|M|} p_m L_m(\theta) = \mathbb{E}_m [L_m(\theta)]$, as defined in \citet{fedprox}. Here, $M$ denotes the set of clients participating in the federated learning (FL) optimization. The term $p_m$ (where $p_m \geq 0$, $\sum_{m=1}^{|M|} p_m = 1$) is an aggregation coefficient, indicating the importance of the learned parameters for client $m$. In most cases, the aggregation coefficient is $p_m=\frac{n_m}{M}$, with $n_m$ being the number of samples trained on client $m$ and $N=\sum_{m=1}^{|M|} n_m$ being the total number of samples in the training dataset. $L_m$ is the local loss function of client $m$, which measures the empirical risk over its local data distribution $D_m$, and is defined as $L_m(\theta) \coloneqq \mathbb{E}_{x_m \sim D_m} [l_m(x_m;\theta)]$.

In some FL applications, the objective is to generate personalized parameters for each client to perform better on their local data while incorporating effective parameter sharing across clients \cite{fallah2020personalized, dinh2022personalized, lee2024fedl2p}. This area within FL is called personalized federated learning (PFL). PFL aims to produce a set of personalized parameters for each client:
$\Theta^* = \textup{arg} \min_{\Theta} \mathbb{E}_m [F_m(\theta_m)]$. Here, $\Theta=\{\theta_m\}_{m=1}^{|M|}$ denotes the set of personalized parameters for all clients.

To tackle the PFL problem, \citet{fallah2020personalized} proposed Per-FedAvg, which adopts a meta-learning framework. pFedMe \cite{dinh2022personalized}, instead, integrates Moreau envelopes \cite{moreau_proprietes_1963} into the clients' regularized loss functions.
SFL \cite{chen2022personalized} utilizes graph-based methods to aggregate parameters following the relational topology among clients.

GHNs and HNs have proven effective in the FL domain due to their parameter-sharing capabilities.
\citet{shamsian_personalized_2021} proposed pFedHN to solve the PFL problem.
\citet{litany2022federated} leveraged GHNs on network computation graphs to tackle the heterogeneity of client model architecture. \citet{xu_heterogeneous_2023} proposed constructing client relation graphs by measuring the cosine similarity between client model parameters and used GHNs to learn relationships between local client data distributions.
\citet{lin_graph-relational_2023} presented Panacea, which couples a GHN with a graph generator to impose a regularization term to preserve the \say{affinity} of client embeddings and maintain their expressiveness.
However, such regularization may be insufficient to prevent over-smoothing from affecting its performance.
Thus, we used cellular sheaves instead of a more standard regularizer.

\section{Methodology: Sheaf HyperNetwork}
\label{subsec:methodology}

\begin{figure}[t]
    \centering
    \includegraphics[width=\linewidth]{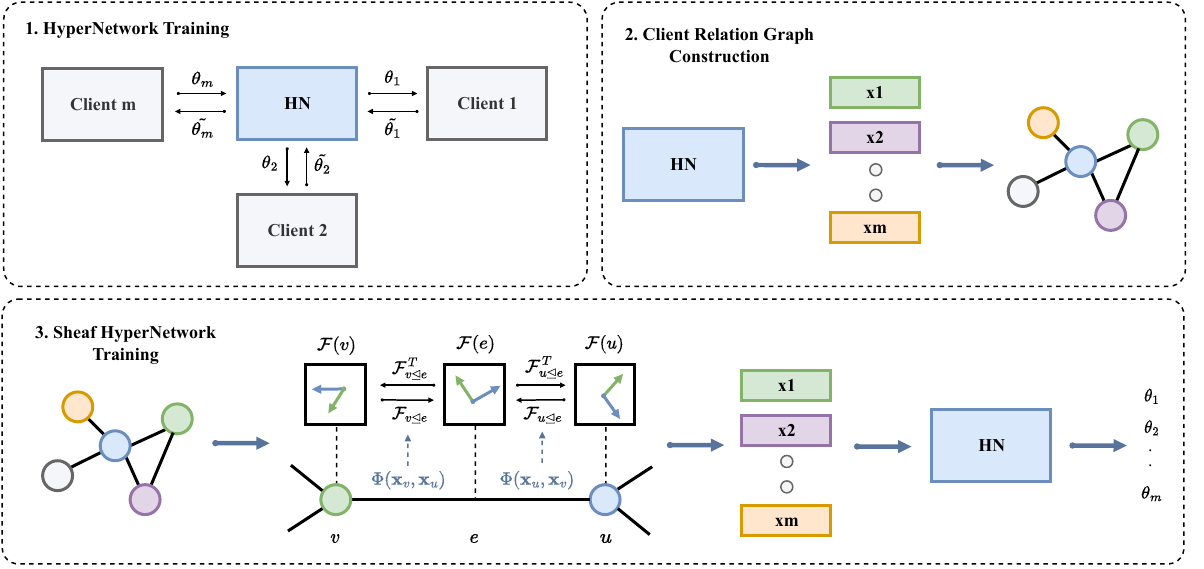}
    \caption{(1) The hypernetwork (HN) is trained to generate personalized parameters for each client collaborating in the personalized federated learning (PFL). (2) The learned client embeddings are extracted from the HN to construct a \emph{client relation graph}. Then, cosine thresholding or $k$-nearest neighbors are used to build edges between similar embeddings. (3) The Sheaf HyperNetwork (SHN) takes on the \emph{client relation graph}. The cellular sheaf projects each client node embedding into the higher-order \emph{stalk} space and uses linear \emph{restriction maps}, learned by the multi-layer perception (MLP) $\Phi$, to guide the diffusion process. After several iterations of sheaf diffusion, rich client embeddings are passed onto an HN to generate personalized parameters for each client.}
    \label{fig:sheaf_hypernetwork_architecture} 
\end{figure}

In this section, we present our novel personalized federated learning (PFL) method, Sheaf HyperNetwork (SHN), to tackle the limitations of existing graph hypernetwork (GHN) solutions. Similarly to other graph-based methods, SHN depends on the client relation graph, which consists of two components: the client node embeddings $X$ and the adjacency matrix $A$. We propose a \emph{three-stage federated training procedure} composed of the following steps.~(1) \textbf{Federated HyperNetwork Training}: we perform PFL with a hypernetwork (HN) to generate personalized parameters for all clients, obtaining the client embeddings $X$.~(2) \textbf{Client Relation Graph Construction}: we extract the learned embedding vectors corresponding to each client and construct a graph based on their cosine similarity, obtaining the adjacency matrix $A$.~(3) \textbf{Federated Sheaf HyperNetwork Training}: we couple a sheaf-based graph neural network with HN and perform PFL training.
Our proposal can seamlessly generalize to any federated learning (FL) setting where $A$ is known, such as traffic \cite{chen2022personalized, lin_graph-relational_2023, meng_cross-node_2021} and weather forecasting \cite{lin_graph-relational_2023}. However, $X$ is unlikely to be known in these settings, so step (1) must be executed. In scenarios where both $X$ and $A$ are available, we execute only step (3), as collecting the information for constructing the client relation graph is superfluous.

\subsection{Federated HyperNetwork Training}
\label{sec:hypernetwork_training}

The primary goal of step (1) is to learn the client embeddings (or node embeddings) $X$ to construct the client relation graph in step (2), as described in \cref{subsec:client_relation_graph_construction}. However, if the embeddings are known, then this step can skipped.
The standard HN's training procedure for PFL aims to generate the optimal personalized model parameters $\theta_m^*$ for each client $m\in M$, where $M$ is the set of clients in the federation.
Each client $m$ has a local optimization objective defined by $\arg \min _{\theta_m} \mathcal{L}_m(\mathcal{D}_m, \theta_m)$, where $\mathcal{D}_m$ is the local client dataset.
Parameterized by $\psi_\mathcal{H}$ and conditioned by $\mathbf{x}_m$, HN generates the personalized model parameters $\theta_m$.
The input $\mathbf{x}_m$ is produced by a multi-layer perception (MLP) $\mathcal{E}$ parametrized by the learnable parameters $\psi_\mathcal{E}$ and conditioned on the client identifier $m$, i.e.~$\theta_m = \mathcal{HN}(\mathbf{x}_m; \psi_\mathcal{H})$ with $\mathbf{x}_m = \mathcal{E}(m; \psi_\mathcal{E})$.
Thus, we can write the HN's optimization objective for PFL as follows:

\begin{equation}
\arg \min _{\psi_\mathcal{H}, \psi_\mathcal{E}} \frac{1}{|M|} \sum_{m=1}^{|M|} \mathcal{L}_m\left(\mathcal{D}_m, \mathcal{HN}\left(\mathcal{E}(m; \psi_\mathcal{E}) ; \psi_\mathcal{H}\right)\right)
\end{equation}

An iterative federated double optimization, described below, carries out the actual minimization procedure for the above objective.
At every federated round $t$, (1) the server selects a subset $S_t$ of clients from the federated population for training.
(2) Then each selected client $m\in S_t$ is sent their personalized model parameters $\theta_m^t = \mathcal{HN}(\mathcal{E}(m, \psi_\mathcal{E}^t), \psi_\mathcal{H}^t)$. (3) Following this, the clients train their models parameterized by $\theta_m^t$ on their respective local datasets $\mathcal{D}_m$. In our experiments, this training uses their local optimizer, e.g.,~stochastic gradient descent (SGD), to minimize their local cost function $\mathcal{L}_m$, e.g.,~cross-entropy, for a classification task. (4) The updated models $\Tilde{\theta_m^t}$ are sent back to the server that executes the server-side optimization. Such optimization consists of minimizing the mean squared error (MSE) between $\Tilde{\theta_m^t}$ and $\theta_m^{t}$, using the Adam \cite{kingma2017adam} optimizer on the parameters $\psi_\mathcal{E}$ and $\psi_\mathcal{H}$. These four steps repeat for a sufficient number of rounds until convergence.
The \emph{client relation graph} construction follows, leveraging the learned MLP $\mathcal{E}$ to generate the clients' embeddings $\mathbf{x}_m=\mathcal{E}(m; \psi_\mathcal{E})$ for each client $m\in M$. Overall, the clients and the server only exchange personalized parameters. We extensively discuss privacy and communication costs in \Cref{appendix:communication_costs,appendix:privacy}.

\subsection{\emph{Client Relation Graph} Construction}
\label{subsec:client_relation_graph_construction}

Constructing a graph where node embeddings $X$ and edges accurately represent FL clients and their relations is challenging because obtaining such information while maintaining privacy is particularly complex. To address this, we propose a new systematic method to construct the \emph{client relation graph} $G=(V, E)$, ensuring that the graph built takes advantage of the relational information available while adhering to privacy-preserving protocols.\footnote{For a client relation graph $G=(V, E)$, each client is represented as a node $v\in V$, and each connection between clients as an edge $e\in E$. We thus have a one-to-one relation between a given HN trained on a federated population and a graph $G$.}

Constructing a \emph{client relation graph} $G$ involves two key steps: (1) Extract $\{\mathbf{x}_v\}_{v\in V}$, each of which defines a \emph{node embedding} of the graph $G$ and characterizes the local data distribution of its respective client. (2) Use the KNN algorithm or cosine thresholding to build the \emph{edges} $e \in E$ of the graph $G$ by linking similar clients based on their embeddings $\{\mathbf{x}_v\}_{v\in V}$.
In rare cases, the \emph{node embeddings} $\{\mathbf{x}_v\}_{v\in V}$ are known a priori, making step (1) redundant.

\textbf{Node Embeddings.}
In the context of PFL, the \emph{node embeddings} within a \emph{client relation graph} aim to encode the clients' local data distributions to facilitate the generation of personalized parameters.
Since the FL paradigm assumes that the clients' data distributions are extremely sensitive, our method learns node embeddings without exchanging private information.
\citet{shamsian_personalized_2021} showed that the clients' embeddings from HN reveal that the clients sharing the same class labels cluster together when trained on a superclass-divided CIFAR100 dataset. 
This evidence supports using learned HN embeddings as descriptive node representations within our \emph{client relation graph}.
As such, our method uses the HN's client embeddings $\mathbf{x}_m=\mathcal{E}(m; \psi_\mathcal{E})$ for each $m\in M$ to construct the clients' node embeddings, as in $\{\mathbf{x}_m\}_{m\in M}=\{\mathbf{x}_v\}_{v\in V}$  with $M\equiv V$, to equip the nodes of the \emph{client relation graph} $G=(V, E)$.

\textbf{Edges.} After extracting client node embeddings from the hypernetwork, we construct the graph's edges $e \in E$ by connecting nodes that exhibit similarity, facilitating effective parameter sharing.
Firstly, we calculate the cosine similarity between all node embeddings pairs, $S_C(\mathbf{x}_v,\mathbf{x}_u)=\frac{\mathbf{x}_v\cdot\mathbf{x}_u}{\parallel\mathbf{x}_v\parallel\parallel\mathbf{x}_u\parallel},~\forall v,u\in V$.
We use the $k$-nearest neighbors (KNN) or cosine thresholding with threshold $\tau$ to establish connections ($e \in E$) based on these similarities.
This process introduces critical hyperparameters, depending on which method is adopted, influencing the graph's underlying topology: (1) the number of neighbors, $k$, to connect nodes, or (2) the similarity threshold $\tau$ required for edge formation.
The edges construction function can be denoted as: $a_{vu} = a_{k/\tau}(\mathbf{x}_v, \mathbf{x}_u)\in\{0, 1\}$ with $v,u\in V$.
The adjacency matrix $A$, whose elements are $a_{vu}$, is a binary, symmetric square matrix with a non-zero diagonal.

\subsection{Federated Sheaf HyperNetwork Training}
\label{sec:sheaf_hypernetwork_training}

Consider a \emph{client relation graph} $G=(V, E)$ constructed as described in \cref{subsec:client_relation_graph_construction}.
Let $n = |V|$ denote the number of nodes.\footnote{The number of nodes is the same as the number of clients, as such $|V|=|M|$}
Each node $v \in V$ has a corresponding feature vector $\textbf{x}_v$ that lies in the row space of $X^{n \times f}$, where $f$ is the node feature dimension. We then
 linearly project $X^{n \times f}$ into $\Tilde{X} \in \mathbb{R}^{n \times (df)}$, by an MLP, which is then reshaped to $\Tilde{X} \in \mathbb{R}^{(nd) \times f}$, whose columns are vectors in $C^0(G ; \mathcal{F})$. This results in each node embedding, represented in the vertex stalk, as a matrix $\mathbb{R}^{d \times f}$, where $d$ is the \emph{stalk dimension}.
As $d$ increases, the cellular sheaf $\mathcal{F}$ adds more information and expressivity to the graph $G$.
However, the choice for $d$ is critical and must be balanced, as an excessive value would result in an overparameterized model, which is more likely to overfit. In our experiments, the same $d$-dimensional space is assigned for all node stalks $\mathcal{F}(v)=\mathbb{R}^d$ and edge stalks $\mathcal{F}(e)=\mathbb{R}^d$. \Cref{eq:sheaf_diffusion} describes the diffusion process over a graph $G$ equipped with the cellular sheaf $\mathcal{F}$.
This is then discretized, leading to the model we incorporate, provided by \citet{bodnar_neural_2023}, given in \Cref{eq:bodnar_sheaf}.
\vspace{-0.14em}

\begin{definition}
    \textup{The \emph{sheaf diffusion process} on ($G, \mathcal{F}$) is a  defined by the differential equation:}
    \begin{equation}
    \Tilde{X}(0)=\Tilde{X}, \frac{\partial}{\partial t} \Tilde{X}(t)=-\Delta_{\mathcal{F}} \Tilde{X}(t)
    \label{eq:sheaf_diffusion}
    \end{equation}
\end{definition}

\begin{definition}
\textup{The model proposed by \citet{bodnar_neural_2023} takes the form:}
\begin{equation}
\Tilde{X}_{t+1}=\Tilde{X}_t-\sigma\left(\Delta_{\mathcal{F}(t)}\left(I \otimes W_1^t\right) \Tilde{X}_t W_2^t\right)     \label{eq:bodnar_sheaf}  
\end{equation}
\textup{where $\sigma$ is a non-linear activation function. The symmetrically normalized sheaf laplacian $\Delta_{\mathcal{F}(t)}$ is defined by the restriction maps, which are learnt by an MLP $\Phi$, where $\mathcal{F}_{v \unlhd e:=(v, u)}=\Phi\left(\Tilde{\mathbf{x}}_v, \Tilde{\mathbf{x}}_u\right) \in \mathbb{R}^{d \times d}$, which we impose to be orthogonal (further discussed in \Cref{appendix:cellular_sheaf_diffusion}). In practice, $\Phi$ takes the concatenation of the two node features $\Tilde{\mathbf{x}}_u$ and $\Tilde{\mathbf{x}}_v$ as input. Then the output is reshaped to map the dimensions of the restriction maps $\mathbb{R}^{d \times d}$. $W_1^t \in \mathbb{R}^{d \times d}$, $W_2^t \in R^{f_{in} \times f_{out}}$ are two time-dependent weight matrices of two MLP layers, implying the underlying \say{geometry} changes over time. $f_{in}$ and $f_{out}$ are the input and output dimensions, and $\otimes$ is the Kronecker product.}
\end{definition}

Following several rounds of the sheaf diffusion process, we obtain a set of rich client node embeddings carrying the information regarding their neighboring client data distributions.
For all clients sampled for training, their corresponding node embedding is selected and passed on to our HN model. Similar to \citet{shamsian_personalized_2021}, our HN consists of a series of MLPs, which takes a node embedding as input to generate the target model's parameters layer-wise. We optimize the sheaf layers and HN with Adam \cite{kingma2017adam}. \Cref{appendix:training_details} describes the training details and the hyper-parameter search space.

\section{Experimental Setup}
\label{sec:experimental_setup}

We conduct experiments across $5$ datasets and $4$ model architectures, encompassing standard FL benchmarks and real-life datasets that naturally conform to graph structures. 
We describe the datasets, model architecture, and federated learning settings in detail below and further in \Cref{appendix:experimental_setup}.

\textbf{Datasets.} 
We evaluate our method using datasets with and without predefined graph topologies to assess its generalizability.
CIFAR100 \cite{krizhevsky_learning_nodate} represents the dataset lacking a predefined topology.
To simulate a realistic cross-device FL environment for CIFAR100, we partition the dataset across a $100$ client pool and induce client heterogeneity using two approaches.
The first method involves Latent Dirichlet Allocation (LDA) \cite{yurochkin_bayesian_2019, qiu_zerofl_2022} to partition data among clients.
The second method, Cluster CIFAR100, assigns $100$ clients to one of the twenty superclasses, similarly to \citet{shamsian_personalized_2021}.
Such partitioning creates a graph with a distinct underlying topology, highlighting the limitations of naive parameter aggregation.
Since parameters across different superclasses reflect diverse input signals, straightforward parameter matching becomes ineffective.
Additionally, we incorporate four other datasets with predefined graph topologies: CompCars \cite{noauthor_compcars_nodate}, TPT-48 \cite{post_washingtonpostdata-2c-beyond--limit-usa_2024}, METR-LA \cite{metr_la_dataset} and PEMS-BAY \cite{pems_bay_dataset}.
We give more details regarding the datasets in \Cref{appendix:datasets,appendix:latent_dirichlet_allocation}.

\textbf{Client Model Architectures.}
For CIFAR100 and Cluster CIFAR100, we use a CNN with two convolutional layers, a max pool layer, and three fully connected layers.
We adhere to the same architectures described by \citet{lin_graph-relational_2023} for the other real-world datasets.
The TPT-48 dataset \cite{post_washingtonpostdata-2c-beyond--limit-usa_2024} leverages a three-layer perceptron.
The traffic and weather forecasting datasets utilize a single-layer Gated Recurrent Unit (GRU) model \cite{cho_learning_2014}, whereas the CompCars dataset \cite{noauthor_compcars_nodate} trains a ResNet-18 model \cite{he2016deep}, pre-trained on the ImageNet \cite{deng_imagenet_2009} dataset.

\textbf{Federated Learning Settings.}
In all experiments, we perform \num{2000} communication rounds, randomly sampling $5$ clients per round.
Each selected client performs local updates using Stochastic Gradient Descent (SGD) for $50$ local optimization steps before transmitting the updated parameters back to the server for aggregation.
We assess \textbf{personalized} performance by instructing each client to evaluate their local model parameters using their respective local test datasets.
We then calculate the mean and standard deviation of the performance across \textbf{all} clients.
We repeated each experimental configuration \num{5} times, initializing the pseudo-random number generators (PRNGs) with different seeds.
Our plots and tables report the average mean and standard deviation.
Further details on the personalized evaluation metric are in \Cref{appendix:evaluation_metric}.

\section{Experimental Results}
\label{sec:experimental_results}

\begin{figure}[t]
    \centering
    \begin{subfigure}{0.26\textwidth}
        \centering
        
        \includegraphics[width=\linewidth]{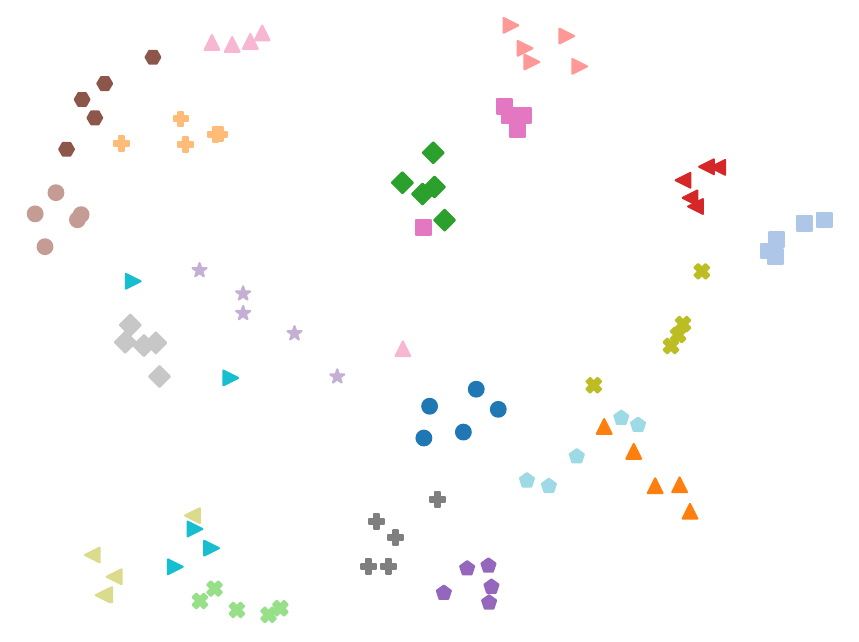}
        
        \caption{}
        \label{fig:cluster_cifar_100_tsne} 
        
    \end{subfigure}\hfill
    \begin{subfigure}{0.23\textwidth}
        \centering
        \includegraphics[width=\linewidth]{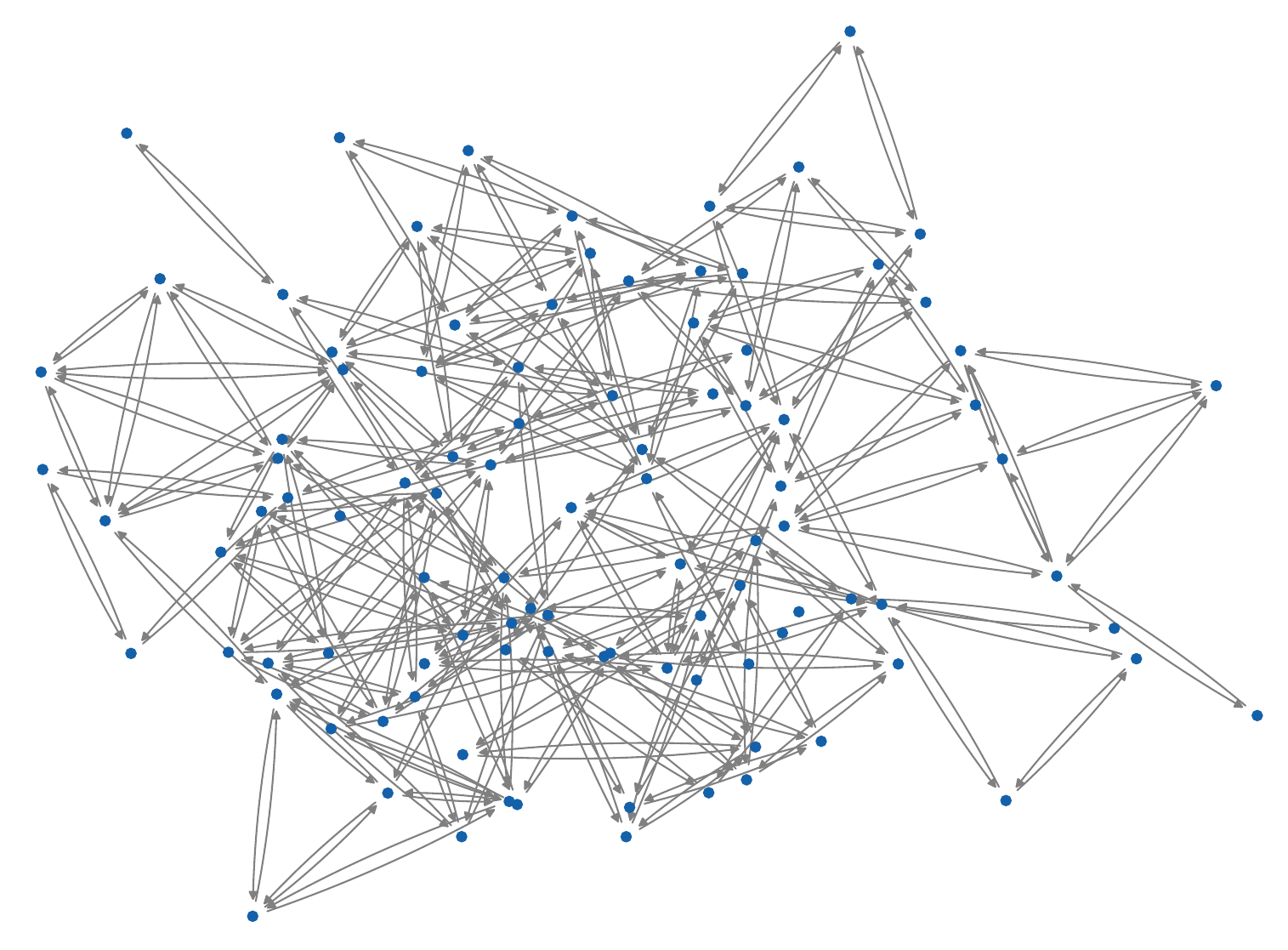}
        \caption{}
        \label{fig:cluster_cifar100_cosine_thresholding_095}
    
    \end{subfigure}\hfill
    \begin{subfigure}{0.23\textwidth}
        \centering
        \includegraphics[width=\linewidth]{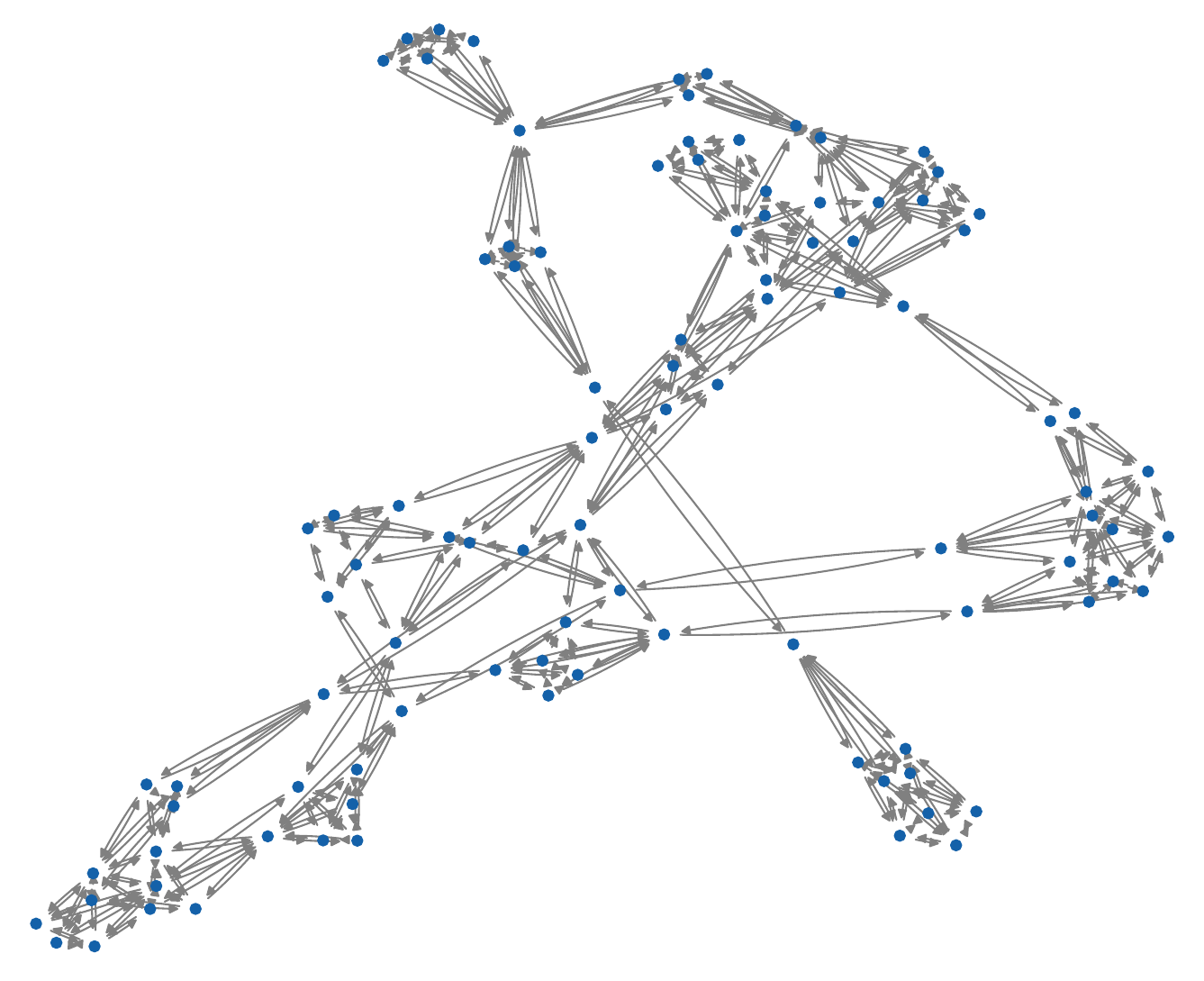}
        \caption{}
        \label{fig:cluster_cifar100_3_knn}
        
    \end{subfigure}\hfill
    \begin{subfigure}{0.23\textwidth}
        \centering
        \includegraphics[width=\linewidth]{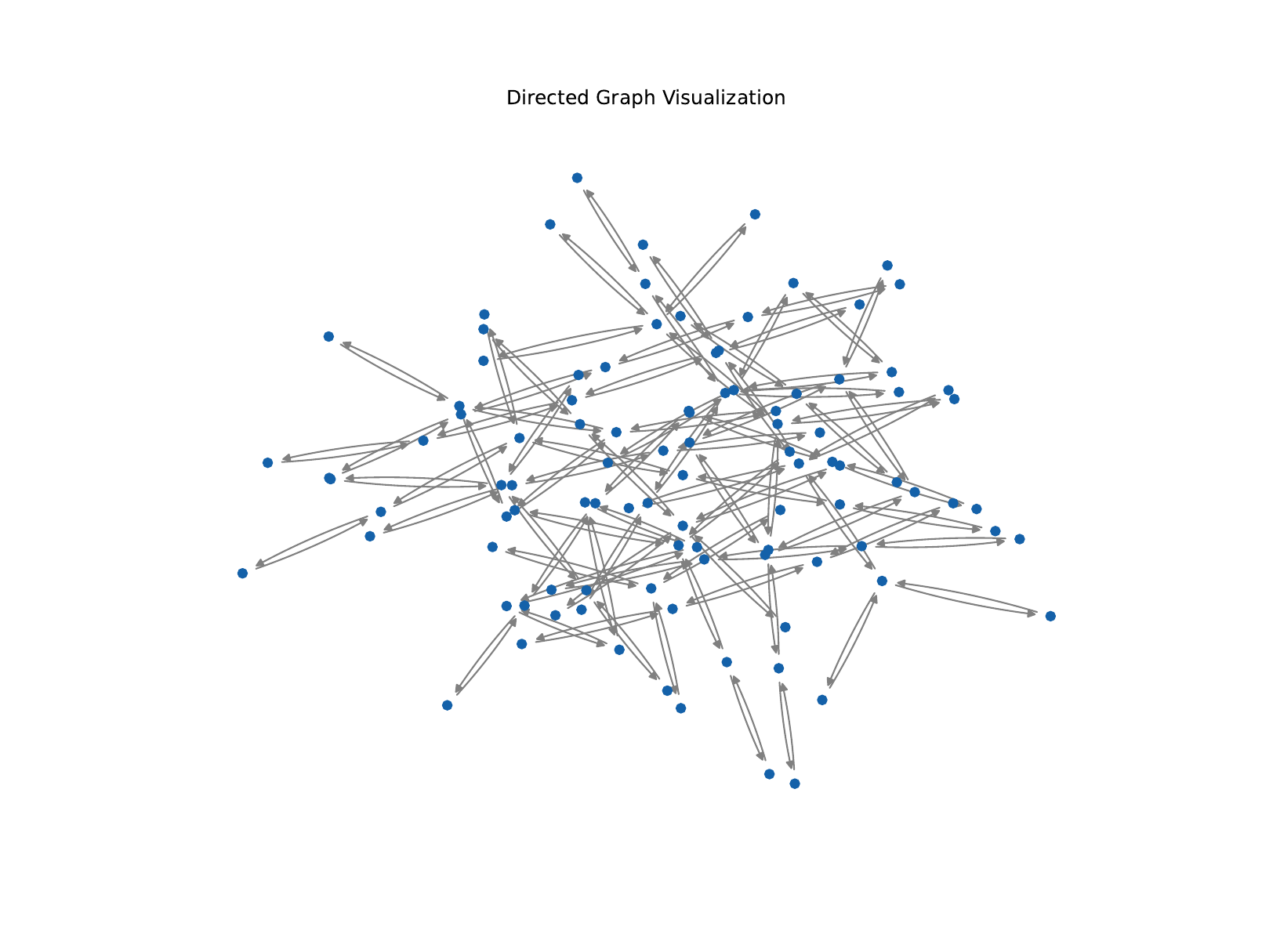}
        \caption{}
        \label{fig:cluster_cifar100_1_knn}

    \end{subfigure}%
    \caption{(a) t-SNE plot of client embeddings for Cluster CIFAR100. Each marker denotes a client belonging to one of the $20$ superclasses. (b) Client relation graph constructed by cosine similarity thresholding, with a value of $0.95$ (c) Client relation graph constructed by KNN with $k=4$ (d) Client relation graph constructed by KNN with $k=1$}
\end{figure}

In this section, we present the results of our evaluation.
We begin with a graphical examination of a hypernetwork's (HN) ability to discern local client data distributions, followed by a discussion on how a client relation graph can be effectively constructed, as detailed in \Cref{subsec:client_relation_graph_construction_res}.
Subsequently, we compare SHN's performance with notable baselines, including pFedHN \cite{shamsian_personalized_2021}, Panacea \cite{lin_graph-relational_2023}, and GHN, a GCN-based graph HN.
Our results show that SHN outperforms these baselines on multiple benchmarks due to the additional expressivity provided by the cellular sheaf.
Moreover, SHN mitigates common GNN challenges such as over-smoothing and heterophily.

\subsection{Client Relation Graph Construction}
\label{subsec:client_relation_graph_construction_res}

We visualize the client relation graph obtained from the Cluster CIFAR100 dataset.
\Cref{fig:cluster_cifar_100_tsne} shows a t-SNE representation of the learned client embedding space after the HN training.
Clients from the same superclass cluster together, which showcases our step (1) successfully extrapolating their underlying relationship. The graph's structural properties depend on the methods used. Using KNN (as in \Cref{fig:cluster_cifar100_3_knn,fig:cluster_cifar100_1_knn}) with $k > 0$ typically ensures a connected graph, whereas cosine thresholding (as in \Cref{fig:cluster_cifar100_cosine_thresholding_095}) with a threshold $\tau > -1.0$ may lead to a disconnected graph.

\subsection{Personalized Federated Learning Results}
\label{subsect:results}

\begin{table}[t]
\caption{\small Personalized average test classification accuracy (CIFAR100, Cluster CIFAR100, CompCars) and MSE (TPT-48, METRA-LA, PEMS-BAY) and standard deviation averaged across all clients. We highlight the top three results as \colorbox{firstplace}{first}, \colorbox{secondplace}{second}, and \colorbox{thirdplace}{third}.}
\label{table:fl_dataset_results}
\centering
\begin{adjustbox}{width=\textwidth}
\begin{tabular}{
  l
  S[table-format=2.2(3)]
  S[table-format=2.2(4)]
  S[table-format=2.2(3)]
  S[table-format=2.2(3)]
  S[table-format=1.2(2)]
  S[table-format=3.2(4)]
  S[table-format=2.2(4)]
}
\toprule
\textbf{Method} & \textbf{CIFAR100 ($\alpha = 10^3$)} & \textbf{CIFAR100 ( $\alpha = 0.1$)} & \textbf{Cluster CIFAR100} & \textbf{CompCars} & \textbf{TPT-48 (10\textsuperscript{-3})} & \textbf{METR-LA (10\textsuperscript{-3})} & \textbf{PEMS-BAY (10\textsuperscript{-3})} \\
\midrule
Local & 4.74 \pm 0.20 & 21.28 \pm 10.14 & 45.64 \pm 8.68 & 85.91 \pm 1.75 & \coloredcell{secondplace}{$2.20 \pm 0.55$} & 126.1 \pm 26.42 & 69.80 \pm 49.14 \\
FedAvg & 27.22 \pm 5.79 & 24.38 \pm 8.10 & 14.49 \pm 9.30 & 87.3 \pm 5.33 & 2.60 \pm 0.90 & 11.80 \pm 5.63 & 9.81 \pm 10.11 \\
FedAvg + finetune & \coloredcell{secondplace}{$29.21 \pm 5.86$} & 28.93 \pm 9.88 & 49.78 \pm 9.89 & \coloredcell{secondplace}{$90.01 \pm 5.81$} & 2.60 \pm 1.00 & 10.85 \pm 5.31 & 9.70 \pm 9.71 \\
\midrule
Per-FedAvg & \coloredcell{firstplace}{$29.81 \pm 6.13$} & 26.51 \pm 9.25 & 32.15 \pm 8.55 & 55.26 \pm 7.6 & 2.60 \pm 0.90 & 86.15 \pm 21.61 & 42.21 \pm 29.70 \\
pFedMe & 24.81 \pm 6.19 & \coloredcell{thirdplace}{$43.97\pm10.57$} & 23.60 \pm 10.20 & 88.06 \pm 4.91 & 2.60 \pm 0.90 & 11.70 \pm 5.51 & 10.06 \pm 10.40 \\
SFL & 15.53 \pm 5.05 & 18.87 \pm 9.03 & 13.54 \pm 15.08 & 86.85 \pm 7.90 & 2.90 \pm 1.00 & 12.19 \pm 5.52 & \coloredcell{thirdplace}{$9.44 \pm 10.02$} \\
pFedHN & 20.85 \pm 5.57 & \coloredcell{secondplace}{$45.52 \pm 10.87$} & \coloredcell{thirdplace}{$51.4 \pm 9.12$} & 83.32 \pm 5.35 & \coloredcell{thirdplace}{$2.30 \pm 0.60$} & \coloredcell{secondplace}{$10.14 \pm 4.81$} & \coloredcell{secondplace}{$8.69 \pm 9.31$} \\
Panacea & 26.13 \pm 7.20 & 35.88 \pm 11.84 & 51.17 \pm 19.69 &  \coloredcell{thirdplace}{$88.23 \pm 6.34$} & 2.60 \pm 0.70 & 10.83 \pm 5.82 & 9.66 \pm 9.43 \\
GHN & 25.86 \pm 7.10 & 43.32 \pm 10.71 & \coloredcell{secondplace}{$55.8 \pm 10.20$} & 88.13 \pm 6.46 & 2.40 \pm 0.80 & \coloredcell{thirdplace}{$10.70 \pm 4.91$} & 9.65 \pm 9.44 \\
\textbf{SHN (ours)} & \coloredcell{thirdplace}{$27.66 \pm 6.91$} & \coloredcell{firstplace}{$48.26 \pm 10.95$} & \coloredcell{firstplace}{$57.44 \pm 9.16$} & \coloredcell{firstplace}{$91.30 \pm 5.76$} & \coloredcell{firstplace}{$2.10 \pm 0.50$} & \coloredcell{firstplace}{$10.02 \pm 4.15$} & \coloredcell{firstplace}{$8.25 \pm 9.21$} \\
\bottomrule
\end{tabular}
\end{adjustbox}
\end{table}

\begin{figure}[t]
    \centering
    \begin{subfigure}{0.49\textwidth}
        \centering
        \includegraphics[width=\linewidth]{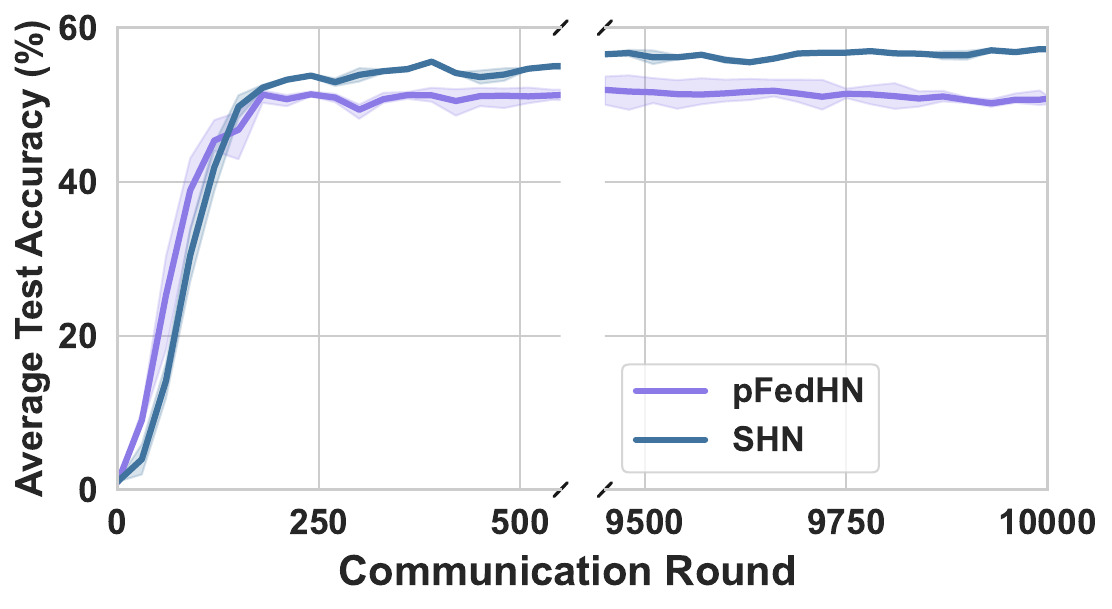}
        \caption{}
        \label{fig:long_running_pFedHN}
    \end{subfigure}\hfill
    \begin{subfigure}{0.48\textwidth}
        \centering
        \includegraphics[width=\linewidth]{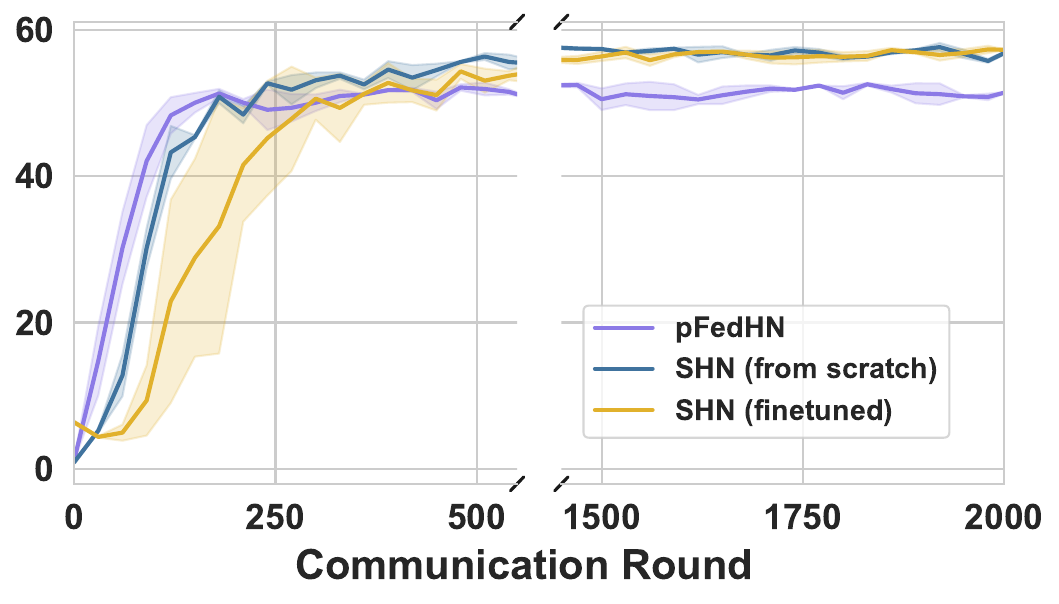}
        \caption{}
        \label{fig:pretrain_finetune_comparison}
    \end{subfigure}
    \caption{\textbf{Cluster CIFAR100:} (a), pFedHN versus SHN for $10$k communication rounds with optimal hyperparameters. (b), SHN (finetuned) initialized from a pre-trained personalized HyperNetwork (pFedHN) against SHN (from scratch) not adopting pFedHN's pre-trained parameters.}
\end{figure}

\textbf{CIFAR100.} We now evaluate the PFL performance on CIFAR100. Since the client relation graph for CIFAR100 is unavailable, SHN and GHN operate on the constructed graphs.
As indicated in \Cref{table:fl_dataset_results}, SHN \emph{outperforms all the other baseline methods} in LDA-partitioned CIFAR100 with $\alpha = 0.1$ (most heterogeneous) and Cluster CIFAR100. However, for IID CIFAR100 ($\alpha = 1000$), where local client data distributions are uniform and IID, graph-based PFL methods, including SHN, appear \emph{less effective} when compared to standard methods like FedAvg or FedAvg combined with local finetuning.
The latter result suggests that the relational inductive bias inherent in graph-based solutions might introduce inefficiencies that can detract from performance in such homogeneous settings.
Furthermore, GCN-based methods such as Panacea and GHN do not consistently surpass pFedHN, possibly due to over-smoothing or the varying degrees of heterophily within the client relation graph.
We further analyze these factors in \Cref{sec:effectiveness_of_sheaves}.

\textbf{Real World Datasets.}
For this set of experiments, our method only needs to compute the node embeddings as a formal graph is given.
As shown in \Cref{table:fl_dataset_results}, SHN \emph{outperforms all existing baselines} across various real-world benchmarks, with notable improvements for METR-LA and PEMS-BAY.
Additionally, SHN maintains a lower personalized standard deviation in test accuracy than most baselines.
While Panacea and GHN, which integrate GCN layers, demonstrate comparable performance, their inability to surpass SHN may highlight limitations in their model expressiveness and the negative impact of over-smoothing.
Conversely, in the temperature forecasting (TPT-48) benchmarks, specific PFL approaches lag behind the local baseline, suggesting that these methods may be ineffective in parameter aggregation.

\textbf{Longer Training Time Evaluation.} We assessed whether pFedHN could perform comparably to SHN if afforded additional training time. We extend the training of pFedHN to $10^4$ communication rounds - five times the number of rounds used for our other experiments.
The result, in \Cref{fig:long_running_pFedHN}, reveals that pFedHN, equipped with its optimal hyper-parameters, performs poorly with extended training.
This figure highlights a fundamental limitation in its ability to capture and utilize relational information, unlike SHN. 

\textbf{Adopting Learned Parameters.} \Cref{fig:pretrain_finetune_comparison} illustrates the learning trajectories of SHN when randomly initialized (not finetuned) against SHN that adopts the learned parameters of a pre-trained pFedHN. Notably, SHN, without adopting the pre-trained pFedHN parameters, converges faster, resulting in fewer communication rounds. Also, the performance of the randomly initialized SHN matches that of the SHN utilizing the pre-trained pFedHN.

\subsection{Effectiveness of Sheaves}
\label{sec:effectiveness_of_sheaves}

\begin{figure}[t]
    \centering
    \begin{subfigure}{0.335\textwidth}
        \centering
        \includegraphics[width=\linewidth]{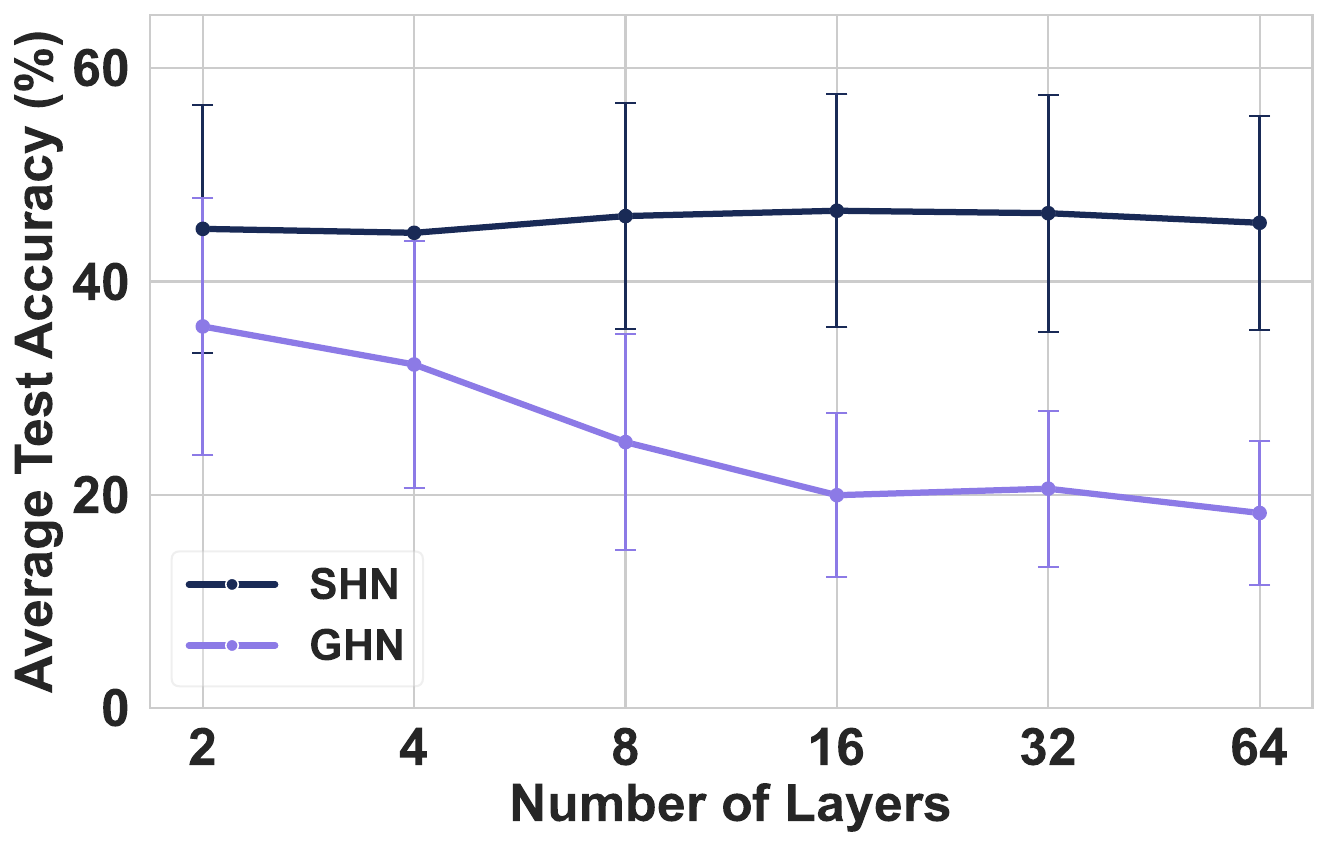}
        \caption{}
    \end{subfigure}
    \hfill
    \begin{subfigure}{0.32\textwidth}
        \centering
        \includegraphics[width=\linewidth]{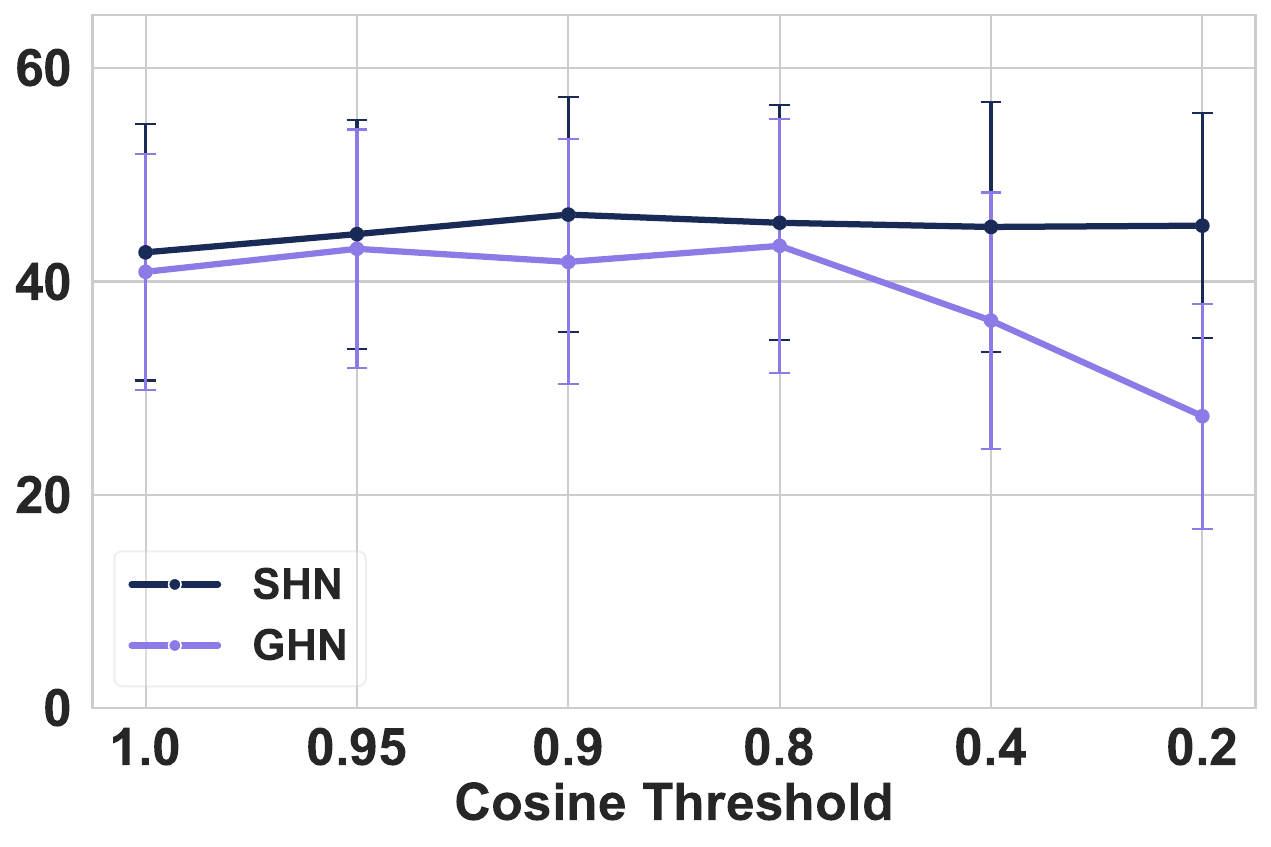}
        \caption{}
    \end{subfigure}
    \hfill
    \begin{subfigure}{0.32\textwidth}
        \centering
        \includegraphics[width=\linewidth]{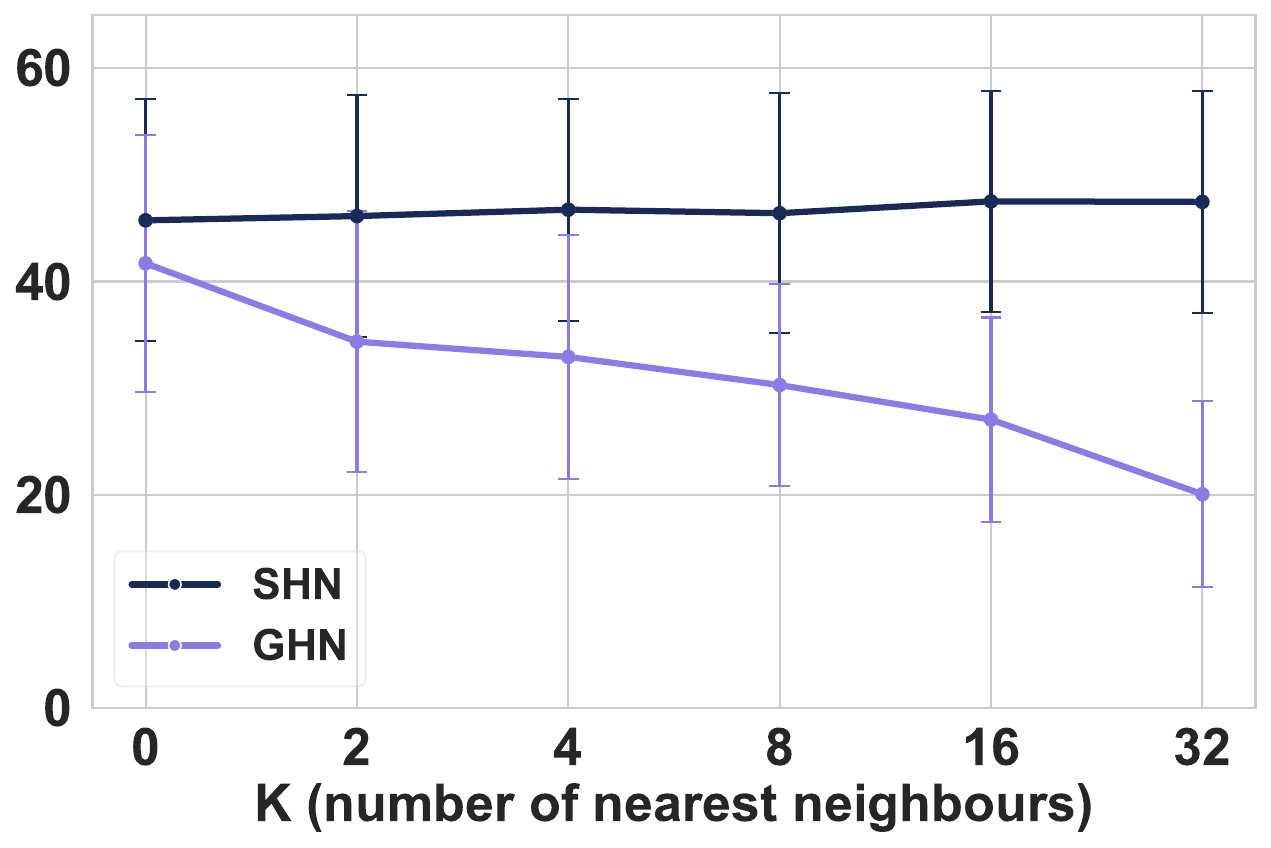}
        \caption{}
    \end{subfigure}
    \caption{Personalized test accuracy \% on CIFAR100.~(a) As the number of message-passing layers increases.~(b) As the cosine threshold decreases.~(c) As the number of $k$-nearest neighbors increases.
    More details can be found in \Cref{appendix:effectiveness_of_sheaves}.
    }
    \label{fig:effectiveness_of_sheaves}
\end{figure}

\textbf{Oversmoothing.}
Previous research \citep{bodnar_neural_2023, duta_sheaf_2023, barbero_sheaf_attention_2022} showed that sheaf neural networks (SNNs) maintain their performance even as the number of message-passing layers increases, a capability that traditional GCN models do not have.
To investigate the over-smoothing problem, we fix the graph topology under the lens of PFL by setting the $k$-nearest neighbors to be $k = 3$ to guarantee a fair comparison between GHN and SHN.
Thus, we isolate the influence of sheaves to gauge their effectiveness better.
We then exponentially increase the number of message-passing layers and assess the performance of SHN against GHN.
As shown in \Cref{fig:effectiveness_of_sheaves}, SHN maintains its performance as the number of layers increases, unlike GHN, whose performance significantly declines after two to four layers.
This decline in GHN is accompanied by a reduction in the standard deviation of the embeddings across layers, illustrating the over-smoothing effect that leads to more homogeneous client embeddings and, consequently, similar performance metrics across all local client datasets.

\textbf{Graph Density Sensitivity.}
In these experiments, we exponentially increase the number of $k$-nearest neighbors and lower the cosine similarity threshold while fixing the number of message-passing layers for GHN and SHN to $3$. Both variations lead to denser graphs, forcing connections that could introduce heterophily and hinder performance.
Even though using denser graphs, \Cref{fig:effectiveness_of_sheaves} shows that SHN is more robust and maintains performance, unlike GHN.
This resilience indicates that SHN can effectively manage non-optimally constructed graphs, a significant advantage in PFL, where extensive hyper-parameter searches could incur high communication costs.

\section{Conclusion}
\label{sec:conclusion}

This work introduces a novel Sheaf HyperNetwork (SHN) model and tackles the issue of data heterogeneity in personalized federated learning (PFL).
Additionally, we propose a privacy-preserving methodology for constructing a client relation graph in settings where it is unavailable.
Such a graph is required for graph-based hypernetwork methods and is essential to leverage underlying client relationships.
Our comprehensive evaluation spans multiple tasks, including multi-class classification, traffic, and weather forecasting, demonstrating SHN's versatility and effectiveness.
Our methodology for constructing a client relation graph introduces new hyperparameters, $k$-nearest-neighbors, and the cosine threshold value, which can be challenging to tune in FL.
However, we show that SHN performs robustly even when such hyperparameters are not optimally set. Overall, SHN outperforms existing PFL baselines by utilizing cellular sheaves to obtain more expressivity and mitigates over-smoothing and heterophily.

\section{Acknowledgments}
\label{sec:acknowledgments}

We would like to thank Federico Barbero (University of Oxford) for his insightful review and constructive suggestions.

\bibliographystyle{unsrtnat}
\bibliography{reference}

\newpage 
\appendix

\section{Appendix}

\subsection{Broader Impact}
\label{appendix:broader_impact}

Federated learning (FL) is a privacy-preserving solution to train machine learning models in a decentralized setting while adopting a parameter-sharing paradigm across collaborating clients. Consequently, previously inaccessible data is now available for machine learning development. From this, FL has gained significant attention and has produced a broad range of applications ranging from health care to preserving patient privacy \cite{noauthor_new_nodate, noauthor_accelerating_2024, rieke_future_2020, flmedicine}, natural language processing \cite{hard_federated_2019, googlefltextselection} to secure finance applications \cite{long_fl_survey, liu2023efficient}. As FL technology is increasingly adopted in academic research and industry, delivering performant solutions to protect personal data is essential. In this work, we present a new class of models called Sheaf HyperNetworks (SHN), which improves upon the current state of the art in an area of FL called personalized federated learning (PFL). Although PFL is within the same umbrella as FL, the optimization objective now shifts towards serving performant bespoke models for each client rather than a single global model. We show SHN is robust against the weaknesses of existing graph-based hypernetwork solutions, such as over-smoothing and heterophily, and offers more expressivity. Combining these three advantages leads SHN to outperform competitive baselines across many domains, such as multi-class classification, weather, and traffic prediction.
Furthermore, our proposed framework for training SHN and constructing a client relation graph does not compromise user privacy. This aspect parallels traditional federated learning methods like FedAvg \cite{mcmahan_communication-efficient_2023}. Consequently, we believe that our research does not contribute to any negative social impacts compared to other personalized federated learning methods. We hope this new class of models will spark further interest within the research community.

\subsection{Future Work}
\label{appendix:future_work}

The proposed sheaf hypernetwork (SHN) model and methodology to construct the client relation graph can be extended in many ways. In this section, we describe potential routes for improving our framework.

\textbf{Client Relation Graph Construction.} When the client relation graph is unknown, our proposed methodology is a three-step process. Instead, this could perhaps be learned. One example is SFL \cite{chen2022personalized}, which constructs the graph using the learned representation of the nodes following several GCN \cite{kipf_semi-supervised_2017} message-passing steps. Our idea is to utilize the opinion dynamics of the cellular sheaf itself. Using the measures of agreement and disagreement, one can learn to construct a homophilic client relation graph throughout the training process.

\textbf{Large Client Models.} In FL, client devices vary from mobile phones to large GPU clusters. Our experiments' largest target client model is ResNet-18, consisting of $11$ million learnable parameters. An interesting study would be to conduct experiments on larger models, on the scale of billions of parameters, in a cross-silo setting and measure the at-the-server computational costs.

\textbf{Client Model Heterogeneity.} In some FL scenarios, each client may hold different model architectures according to their local objective and data. Hypernetworks are abstract and can accommodate this. Future work in this area could provide a more extensive evaluation of SHNs' ability to accommodate architectural heterogeneity. Perhaps this study would be more aligned with the works of \citet{litany2022federated}. As previously mentioned, SHN can be applied to any objective outside of FL. Thus, the works of \citet{zhang_graph_2020}, which utilizes a graph hypernetwork, could perhaps be extended to determine performance improvements in neural architecture search.

\textbf{Global Model.} In this paper, our application of SHN is under a specific umbrella of FL: personalized federated learning. A question arises: Could the provided framework be adjusted to create a single global model?

\subsection{Methodology}
\label{appendix:methodology}

In this section, we provide further details of our methodology. In \Cref{appendix:hypernetwork_training}, we detail a general personalized federated learning pipeline for any hypernetwork. Subsequently, in \Cref{appendix:cellular_sheaf_diffusion,appendix:oversmoothing}, we define the components of a cellular sheaf required for diffusion and theoretically demonstrate how cellular sheaves mitigate over-smoothing. In \Cref{appendix:privacy} and \Cref{appendix:communication_costs}, we discuss our proposed method's privacy-preserving and communication cost aspects. Finally, in \Cref{appendix:adopting_new_clients}, we outline the process for integrating new clients into the system using our approach.

\subsubsection{HyperNetwork Training for Personalized Federated Learning}
\label{appendix:hypernetwork_training}

\begin{algorithm}[t]
    \small
    \DontPrintSemicolon
    \KwIn{\Big[\colorbox{firstplace}{$\mathcal{HN}$}, \colorbox{thirdplace}{$\mathcal{GHN}$}, \colorbox{secondplace}{$\mathcal{SHN}$}\Big] Number of communication rounds - $R$, learning rate - $\eta > 0$}
    \KwIn{\Big[\colorbox{thirdplace}{$\mathcal{GHN}$}, \colorbox{secondplace}{$\mathcal{SHN}$}\Big] Client relation graph, adjacency matrix and embeddings: $G$, $A$, $X^*$.}
    \KwIn{\Big[\colorbox{secondplace}{$\mathcal{SHN}$}\Big] \emph{stalk} dimension - $d$}
    \KwOut{Parameters of Hypernetwork.}
    \BlankLine
    \For{\textup{each communication round} $t=1, 2 .. ,R$}{
        $\Theta_{pred},  \Theta_{opt} \gets [ \ \ ], [ \ \ ]$\;
        $S_{t} \gets$ sample subset of clients from $M$\;

        \uIf{not X}{$X \gets \mathcal{E}(M; \psi_{\mathcal{E}})$\;}
        
        \For{\textup{client} $m \in S_{t}$ \textup{in parallel}}{
            \colorbox{firstplace}{\parbox{0.65\linewidth}{$\textbf{x}_m \gets$ GetClientEmbedding($m$, $X$)\;
            $\theta_{m} \gets \mathcal{HN}(\textbf{x}_{m}; \psi_{\mathcal{H}})$}}\;
            \colorbox{thirdplace}{\parbox{0.65\linewidth}{$\theta_{m} \gets \mathcal{GHN}(m, G, A, X; \psi_{\mathcal{H}})$}}\;
            \colorbox{secondplace}{\parbox{0.65\linewidth}{$\theta_{m} \gets \mathcal{SHN}(m, d, G, A, X; \psi_{\mathcal{H}})$}}\;
            $\Tilde{\theta_{m}} \gets$ ClientUpdate($m$, $\theta_{m}$)\;
            $\textup{Append}(\Theta_{opt},\Tilde{\theta_{m}})$\;
            $\textup{Append}(\Theta_{pred}, \theta_{m})$\;
        }
   
        $\psi_{\mathcal{H}} \gets \psi_{\mathcal{H}} - \eta\nabla_{\psi_{\mathcal{H}}} \mathcal{L_{H}} (\Theta_{pred}, \Theta_{opt})$\;
        
        \uIf{not X}{$\psi_{\mathcal{E}} \gets \psi_{\mathcal{E}} - \eta\nabla_{\psi_{\mathcal{E}}} \mathcal{L_{H}} (\Theta_{pred}, \Theta_{opt})$\;}
        
    }
    \Return{$\psi_{\mathcal{H}}, \psi_{\mathcal{E}}$}
    \caption{\small Hypernetwork Training for Personalized FL}
    \label{alg:train_hypernetwork}
    
\end{algorithm}

\Cref{alg:train_hypernetwork} details a generic hypernetwork pipeline for personalized federated learning (PFL). All hypernetwork models depend on two hyperparameters: the number of federated communication rounds $R$ and its learning rate $\eta$. If a graph-based hypernetwork, such as GHN and SHN, is chosen, then the client relation graph $G$, adjacency matrix $A$, and, optionally, the client node embeddings $X$ are provided. However, when using SHN, the \emph{stalk} dimension $d$ must also be provided. 

With the necessary inputs given, the federated process begins. A fixed number of clients is randomly sampled at the server for training for each communication round. Following this, if the client node embeddings $X$ are not provided as input, the multi-layer perceptron MLP $\mathcal{E}$ parameterized by $\psi_{\mathcal{E}}$ generates the embeddings for all clients in $m$. Importantly, the $\mathcal{E}$ is only conditioned by a scalar client identifier $m$. Subsequently, for each client $m$ in parallel in the case of \big(\colorbox{firstplace}{$\mathcal{HN}$}\big), the client embedding corresponding to the selected client is extracted from $X$. Then, the hypernetwork $\mathcal{HN}$ generates personalized parameters $\theta_m$. \big(\colorbox{thirdplace}{$\mathcal{GHN}$}\big) takes on all the elements of the client relation graph and performs message passing through GCN \cite{kipf_semi-supervised_2017} message-passing layers. \big(\colorbox{secondplace}{$\mathcal{SHN}$}\big) also takes on all the elements of the client relation graph and the stalk dimension $d$. Subsequently, message-passing is performed via the cellular sheaf diffusion process \Cref{eq:bodnar_sheaf}.

\textbf{Remark.} Notably, the graph-based methods GHN and SHN operate on all client node embeddings. In contrast, HN operates on a single embedding belonging to the client selected for training. Therefore, HN acts on a one-to-one correspondence between the embeddings $\textbf{x}_m$ and the personalized parameters $\theta_m$. The graph-based methods GHN and SHN operate across the graph and thus inject a relational inductive bias into the system. This bias allows us to leverage the hidden topology defined by the client relation graph to generate more performant parameters.

Following the generation of the personalized parameters $\theta_m$ from the hypernetwork, the parameters are sent to the corresponding client $m$ for local optimization. \Cref{alg:client_update} describes the client update process locally on each device. The client $m$ receives its personalized parameters $\theta_{m}$ and optimizes it for several epochs on its local data. Subsequently, the updated parameters are sent to the server.

The server receives the optimized client embeddings $\Tilde{\theta}_m$. Both $\Tilde{\theta}_m$ and $\theta_m$ are appended to the lists $\Theta_{opt}$ and $\Theta_{pred}$, respectively. The communication round is complete after every randomly selected client has performed this process. Finally, the parameters $\psi_{\mathcal{H}}$ and $\psi_{\mathcal{E}}$ are updated using the loss function $\mathcal{L_{H}}$, which takes as input the list of predicted and updated parameters. $\mathcal{L_{H}}$ is the mean squared error loss in our experiments.

\begin{algorithm}[H]
    \DontPrintSemicolon
    \KwIn{Client $m$, Initial client model parameters $\theta_m$}
    \KwOut{Updated client model parameters $\Tilde{\theta}_m$}
    \BlankLine
    $\Tilde{\theta}_m \gets \theta_m$\;
    Load local data $\mathcal{D}_m$\;
    \For{\textup{each local epoch} $e = 1, 2, \ldots, E$}{
        \For{\textup{each batch} $b \in \mathcal{D}_m$}{
            Update $\Tilde{\theta}_m$ using gradient descent: $\Tilde{\theta}_m \gets \Tilde{\theta}_m - \eta \nabla \mathcal{L}_{m}(\Tilde{\theta}_m; b)$\;
        }
    }
    \Return{$\Tilde{\theta}_m$}
    \caption{ClientUpdate: Ran on client-side}
    \label{alg:client_update}
\end{algorithm}

\subsubsection{Cellular Sheaf Diffusion}
\label{appendix:cellular_sheaf_diffusion}

As defined in \Cref{def:cellular_sheaf}, the restriction maps $\mathcal{F}_{v \unlhd e}$ facilitate the flow of information from the stalks $\mathcal{F}(v) \rightarrow \mathcal{F}(u)$, where $v, u \in V$ are two nodes in the graph. These restriction maps are learned by an MLP $\Phi: \mathbb{R}^{d \times 2} \rightarrow \mathbb{R}^{d \times d}$ such that $\mathcal{F}_{v \unlhd e:=(v, u)}=\Phi\left(\Tilde{\mathbf{x}}_v, \Tilde{\mathbf{x}}_u\right)$. As described in \Cref{sec:sheaf_hypernetwork_training}, the two node features $\Tilde{\mathbf{x}}_v$ and $\Tilde{\mathbf{x}}_u$ are concatenated and mapped to the dimensions of the restriction maps, i.e.,~$\Phi\left(\Tilde{\mathbf{x}}_v, \Tilde{\mathbf{x}}_u\right)=\sigma\left(\mathbf{W}\left[\text{vec}(\Tilde{X}_u) \| \text{vec}(\Tilde{X}_v)\right]\right)$. Where $\text{vec}(\cdot)$ converts a $d \times f$ matrix into a vector $\mathbb{R}^{df}$ i.e. $\text{vec}(\cdot): \mathbb{R}^{d \times f} \rightarrow \mathbb{R}^{df}$. Following this, constraints can be imposed on the class of restriction maps to allow for different separation properties. These classes include \emph{general linear}, \emph{diagonal}, and \emph{orthogonal}. The \emph{general linear} restriction map is the most powerful but prone to overfitting and is harder to train due to being unbounded. \emph{Diagonal} restriction maps mitigate this by reducing the number of learnable parameters, but the stalk dimensions cannot mix as the non-diagonal terms are $0$. \emph{Orthogonal} restriction maps balance efficiency and generality. Furthermore, since the eigenvalues of the orthogonal maps are $\pm1$, having multiple layers does not result in exploding or vanishing gradients. Thus, it is for this reason that our restriction maps are orthogonal.

\textbf{Remark.} Motivated by Riemannian geometry, \citet{barbero_sheaf_2022} propose to compute these cellular sheaves with orthogonal restriction maps via Principle Component Analysis. This process can be as performant and carries less computational overhead compared to learning these restriction maps with the MLP $\Phi$. However, this requires knowledge of the node embeddings beforehand, which is not always the case in FL. Therefore, we must learn the cellular sheaves.

\begin{definition}
The space of $0$-cochains and $1$-cochains is defined by the sum over the node stalks and edge stalks, respectively: $C^0(G, \mathcal{F}):=\bigoplus_{v \in V} \mathcal{F}(v)$, $C^1(G, \mathcal{F}):=\bigoplus_{e \in E} \mathcal{F}(e)$.
\label{def:cochains}
\end{definition}

Following this, a linear co-boundary map can then be constructed $\delta: C^0(G, \mathcal{F}) \rightarrow C^1(G, \mathcal{F})$ which allows us to map the space of $0$-cochains to $1$-cochains. In opinion dynamics, $\delta$ can be seen as a measure of \say{disagreement} between nodes.

\begin{definition}
    Given an arbitrary orientation, for each edge $e = u \rightarrow v$, the linear co-boundary map $\delta: C^0(G, \mathcal{F}) \rightarrow C^1(G, \mathcal{F})$ is defined as: $(\delta x)_e=\mathcal{F}_{v \unlhd e} x_v - \mathcal{F}_{u \unlhd e} x_u$. Here $\mathbf{x} \in C^0(G, \mathcal{F})$ is a $0$-cochain and $\mathbf{x}_v \in \mathcal{F}(v)$ is a vector of $x$ at the node stalk $\mathcal{F}(v)$.
\label{eq:co_boundary_map}
\end{definition}

Having defined $\delta$, we can now construct the sheaf Laplacian operator of a sheaf, allowing us to perform message-passing over the graph.

\begin{definition}
    The sheaf Laplacian of a sheaf is a map $\mathbf{L}_{\mathcal{F}}: C^0(G, \mathcal{F}) \rightarrow C^0(G, \mathcal{F})$, which is defined as: $\mathbf{L}_{\mathcal{F}} = \delta^T\delta$.
\end{definition}

The sheaf Laplacian is a symmetric positive semi-definite block matrix. Whereby the diagonal blocks are $L_{\mathcal{F}_{v, v}}=\sum_{v \triangleleft e} \mathcal{F}_{v \triangleleft e}^{\top} \mathcal{F}_{v \unlhd e}$ and the off-diagonal block are $L_{\mathcal{F}_{v, u}}=-\mathcal{F}_{v \unlhd e}^{\top} \mathcal{F}_{u \unlhd e}$. Subsequently, we can perform symmetric normalization on the matrix so nodes with large degrees do not dominate the matrix properties.

\begin{definition}
    \textup{The \emph{normalized} sheaf Laplacian $\Delta_{\mathcal{F}}$ is defined as $\Delta_{\mathcal{F}}=D^{-\frac{1}{2}} L_{\mathcal{F}} D^{-\frac{1}{2}}$ where $D$ is the block diagonal of $L_{\mathcal{F}}$.}
\label{def:normalised_sheaf_laplacian}
\end{definition}

The sheaf Laplacian generalizes the graph Laplacian, which is defined as $L = D - A$, where $D$ is the degree matrix and $A$ is the graph's adjacency matrix. The graph Laplacian is considered a trivial sheaf, equivalent to setting the stalk dimension to $d = 1$ and the \emph{restriction maps} to identity functions. The sheaf Laplacian allows for added expressivity, equipping the graph with more information through the $stalk$ space when $d \geq 1$.

\subsubsection{Oversmoothing \& Sheaves}
\label{appendix:oversmoothing}

We can examine the over-smoothing effect by using the Dirichlet energy to determine the expressiveness of the node embeddings in the graph. \citet{cai_note_2020} perform this experiment to prove that the Dirichlet energy decreases exponentially to the number of message-passing layers.

\begin{definition}
\textup{The Dirichlet energy $E(x)$ of a scalar function $x \in \mathbb{R}^{|V| \times 1}$ on the weighted graph $G$ is defined as:}

\begin{equation}
\label{eq:dirichlet_energy_weighted_graph}
E(x)=x^T \title{\Delta_G} x=\frac{1}{2} \Sigma w_{u v}\left(\frac{x_u}{\sqrt{1+d_u}}-\frac{x_v}{\sqrt{1+d_v}}\right)^2
\end{equation}

\textup{Here, $\tilde{\Delta_G}$ refers to the augmented normalized Laplacian of $G$. $w_{uv}$ refers to as the weighted edge between nodes $u$ and $v$. $x_u, x_v \in \mathbb{R}$ are the scalar values at position $u$ and $v$ respectively. $d_u, d_v \in \mathbb{Z}$ are the degrees of nodes $u$ and $v$ respectively.}

\textup{Subsequently, the Dirichlet energy over a vector field $X \in \mathbb{R}^{|V| \times f}$ is defined as:}

\begin{equation}
\label{eq:dirichlet_energy_vector_field}
E(X)=\operatorname{tr}\left(X^T \tilde{\Delta} X\right)=\frac{1}{2} \Sigma w_{u v}\left\|\frac{\textbf{x}_u}{\sqrt{1+d_u}}-\frac{\textbf{x}_v}{\sqrt{1+d_v}}\right\|_2^2
\end{equation}

\end{definition}

We can see that minimizing \Cref{eq:dirichlet_energy_weighted_graph} and \Cref{eq:dirichlet_energy_vector_field} causes the node embeddings $h$ to converge to a homogeneous state. Sheaves, however, naturally prevent over-smoothing. We can see this by inspecting the sheaf Dirichlet energy.

\begin{definition}
\textup{The sheaf Dirichlet energy $E_{\mathcal{F}}(\mathbf{x})$, for which sheaf diffusion minimizes is defined by:}
\begin{equation}
\label{eq:sheaf_dirichlet_energy}
E_{\mathcal{F}}(\mathbf{x}):=\mathbf{x}^{\top} \Delta_{\mathcal{F}} \mathbf{x}=\frac{1}{2} \sum_{e:=(u, v)}\left\|\mathcal{F}_{u \unlhd e} D_u^{-1 / 2} \mathbf{x}_u - \mathcal{F}_{v \unlhd e} D_v^{-1 / 2} \mathbf{x}_v\right\|_2^2
\end{equation}
\end{definition}

We can see in \Cref{eq:sheaf_dirichlet_energy}, the minimization occurs between the node representations and the edge stalks, rather than between the node representations as described \Cref{eq:dirichlet_energy_vector_field}. Due to sheaves operating in this way, the over-smoothing problem is mitigated.

\subsubsection{Privacy}
\label{appendix:privacy}

Our framework only requires exchanging personalized parameters between the client and the server. No additional information is needed from the client to construct the client relation graph or to train the hypernetwork. Although we do not empirically measure the similarity of personalized parameters between comparable clients in our experiments, our framework differs from other graph-based methods \cite{xu_graph-relational_2023, baek_personalized_2023} in that it does not perform direct parameter aggregation. Instead, the hypernetwork generates the personalized parameters. Consequently, a client cannot reconstruct the parameters of a neighboring client unless they share the same embedding, which would imply identical local dataset distributions.

\subsubsection{Communication Costs}
\label{appendix:communication_costs}

All hypernetworks in our framework reside on the server.
The server and clients exchange only personalized parameters.
As a result, training the hypernetworks, such as SHN, and constructing the client relation graph do not incur additional communication costs compared to traditional federated learning methods like FedAvg \cite{mcmahan_communication-efficient_2023}.
\vspace{-1em}

\subsubsection{Adopting New Clients}
\label{appendix:adopting_new_clients}

The primary focus of this paper is to demonstrate SHN's expressivity and resistance to over-smoothing and heterophily compared to GHN and other PFL methods. Consequently, we defer experiments on adopting new clients to future work. However, one possible method for incorporating new clients involves the following steps: ($1$) Add the new client to the training pool. ($2$) Train the hypernetwork (HN) to estimate the personalized parameters of the new client. ($3$) Extract the learned client embedding from the HN for that client. ($4$) Reconstruct the client relation graph with the new client embedding. ($5$) Continue PFL with SHN. If the client embedding is already known, one can start from step $4$.
\vspace{-1em}

\subsection{Results \& Discussion}
\label{appendix:results_and_discussion}

In this section, we provide further details on the results of our experiments. In \Cref{appendix:client_relation_graph_construction}, we illustrate the t-SNE plot of client embeddings for LDA-partitioned CIFAR100 under different levels of heterogeneity. \Cref{appendix:federated_learning_results} discusses various aspects of our federated learning results, including the convergence rate of SHN, results from longer-running experiments, and experiments where SHN adopts the learned parameters of HN. Additionally, \Cref{appendix:effectiveness_of_sheaves} delves deeper into the effectiveness of sheaves experiments in \Cref{sec:effectiveness_of_sheaves}, demonstrating SHN's robustness to over-smoothing and heterophily exhibited in the client relation graph.
\vspace{-0.5em}

\subsubsection{Client Relation Graph Construction}
\label{appendix:client_relation_graph_construction}
\vspace{-1em}

\begin{figure}[H]
    \centering
    \includegraphics[width=\linewidth]{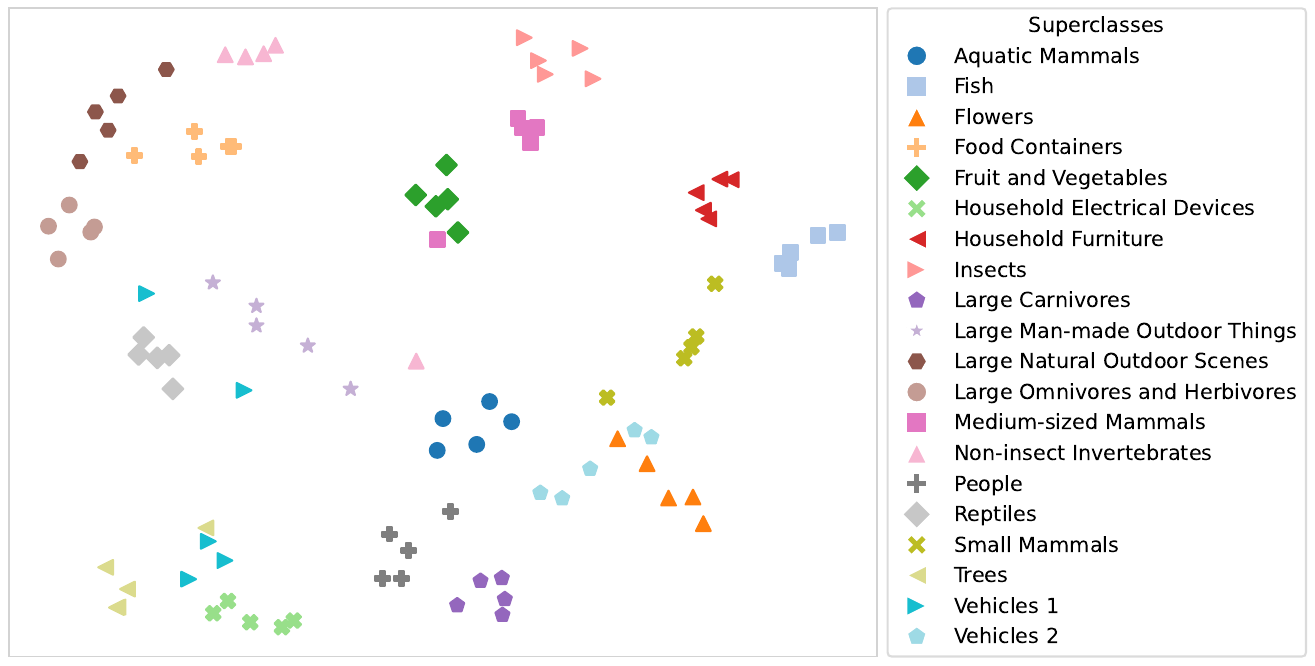}
    \caption{t-SNE plot of the learnt client embeddings for Cluster CIFAR100. Each marker denotes a client belonging to one of the twenty CIFAR100 superclasses.}
    \label{appendix:fig:cluster_cifar_100_tsne} 
\end{figure}

\begin{figure}[H]
    \centering
    \begin{subfigure}{0.24\textwidth}
        \centering
        \includegraphics[width=\linewidth]{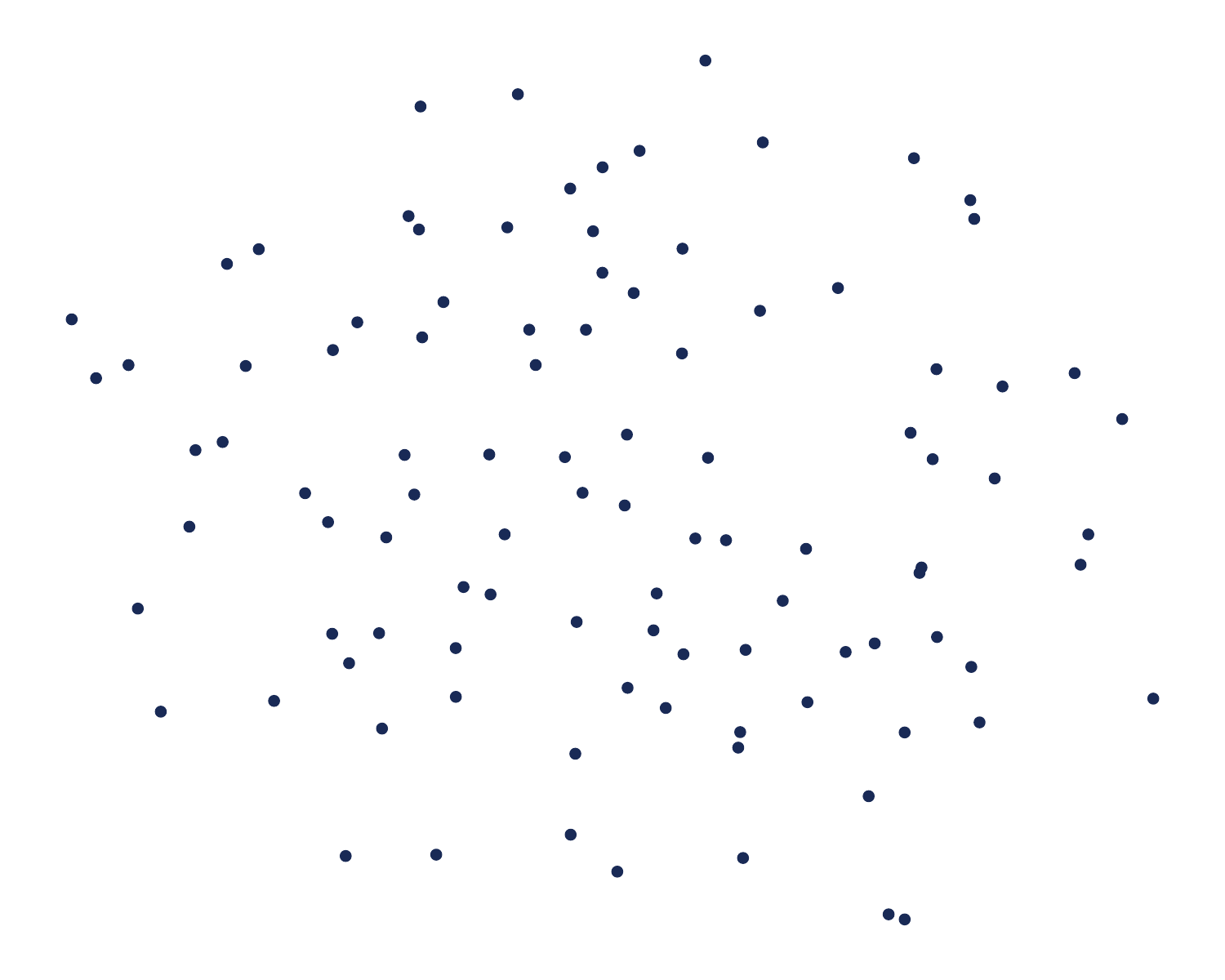}
        \caption{$\alpha = 10^{15}$}
    \end{subfigure}
    \hfill
    \begin{subfigure}{0.24\textwidth}
        \centering
        \includegraphics[width=\linewidth]{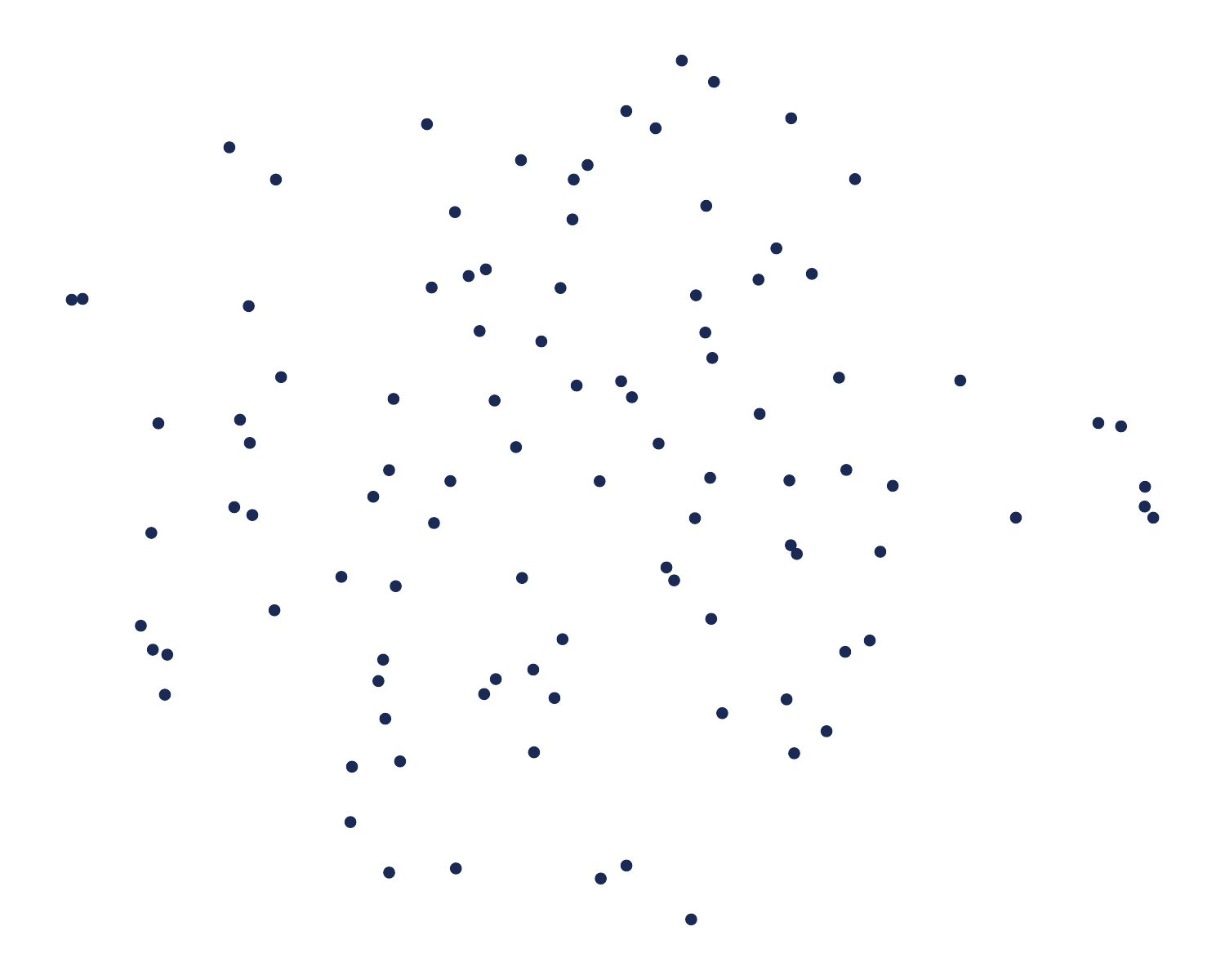}
        \caption{$\alpha = 100.0$}
    \end{subfigure}
    \hfill
    \begin{subfigure}{0.24\textwidth}
        \centering
        \includegraphics[width=\linewidth]{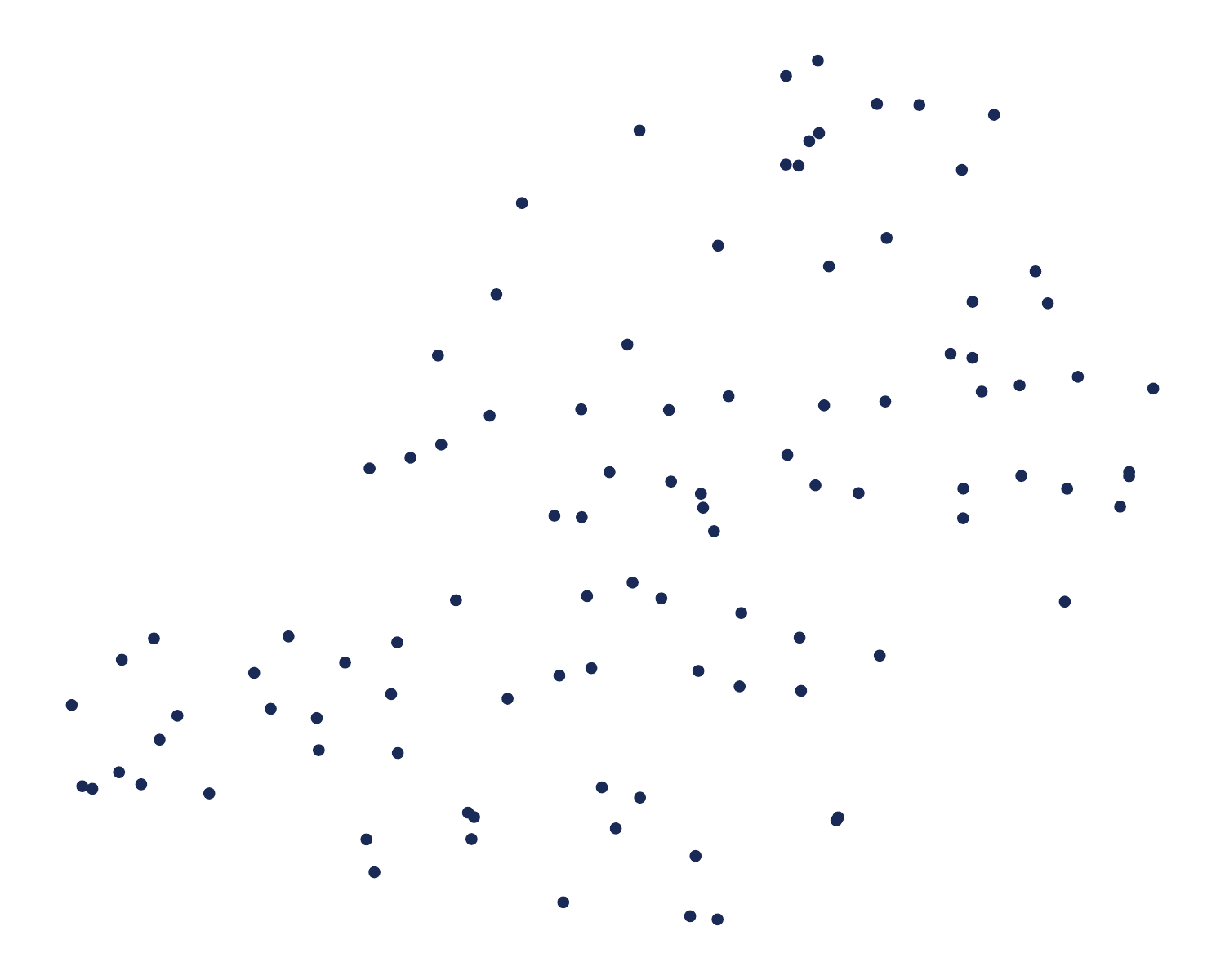}
        \caption{$\alpha = 0.1$}
    \end{subfigure}
    \hfill
    \begin{subfigure}{0.24\textwidth}
        \centering
        \includegraphics[width=\linewidth]{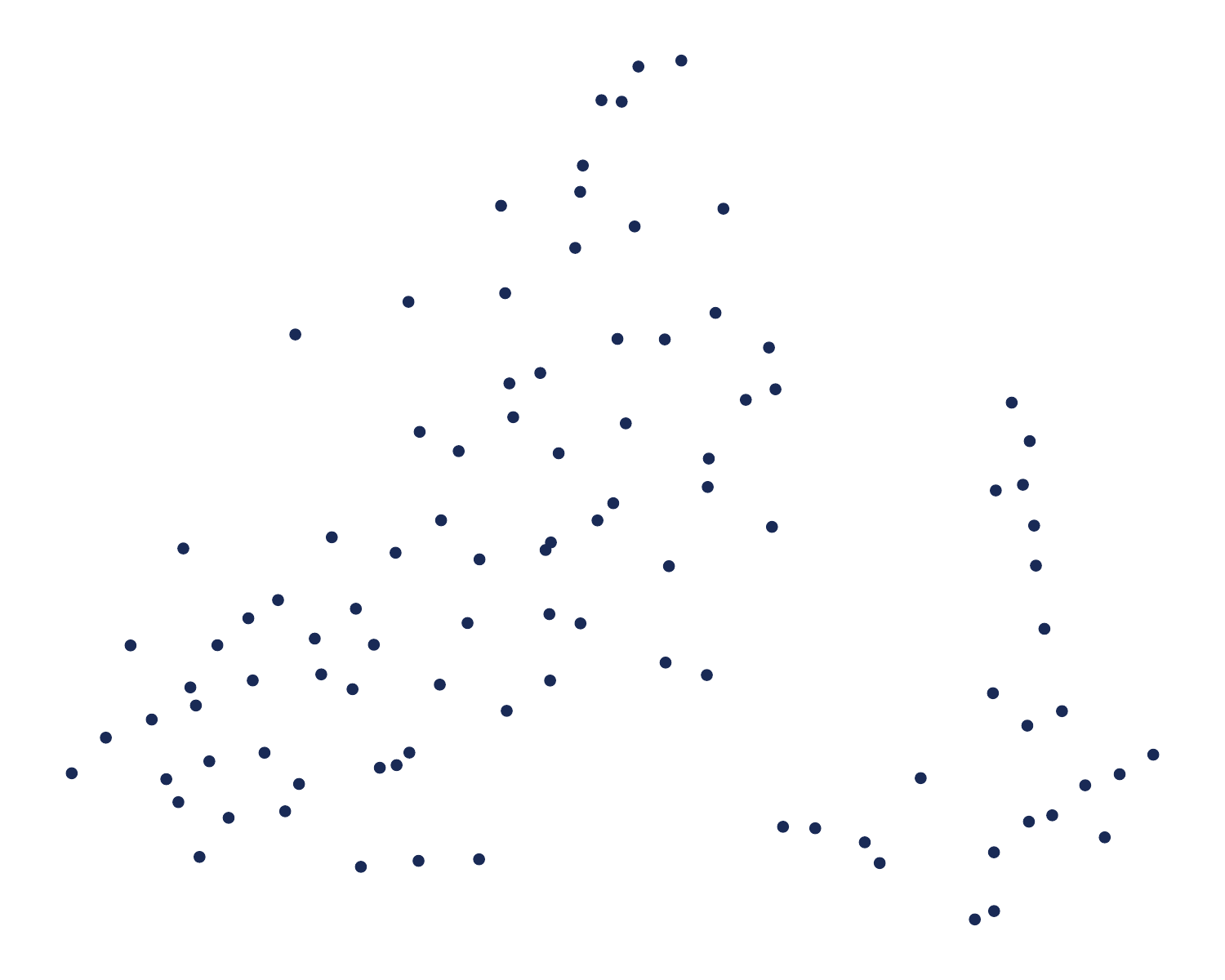}
        \caption{$\alpha = 10^{-15}$}
    \end{subfigure}
    \caption{t-SNE embeddings from personalized hypernetwork, trained on LDA partitioned CIFAR100 with differing $\alpha$ values.}
    \label{fig:tsne_lda_cifar100}
    \vspace{-1.5em}
\end{figure}

\begin{figure}[H]
    \centering
    \begin{subfigure}{0.24\textwidth}
        \centering
        \includegraphics[width=\linewidth]{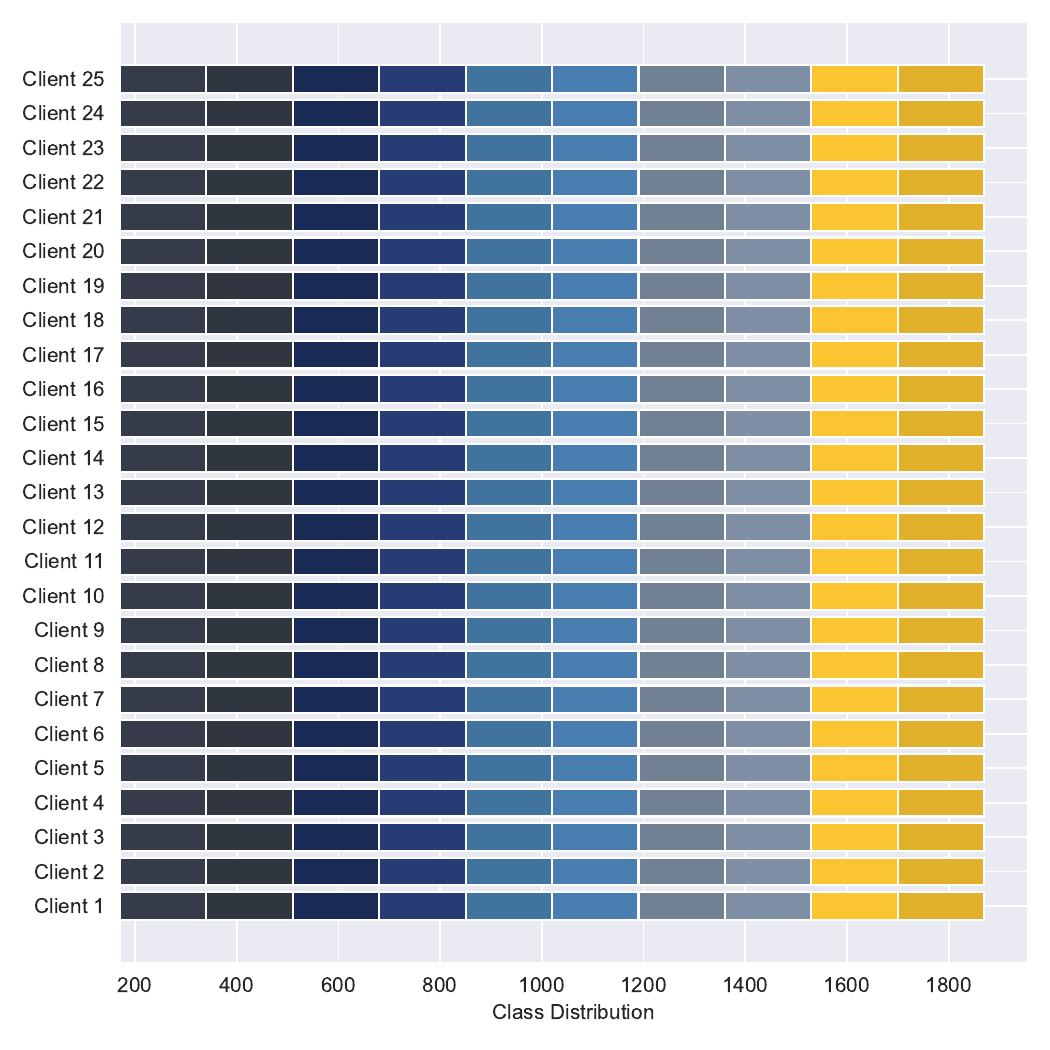}
        \caption{$\alpha \rightarrow \infty$}
        \label{app:fig:lda_inf}
    \end{subfigure}%
    \hfill
    \begin{subfigure}{0.24\textwidth}
        \centering
        \includegraphics[width=\linewidth]{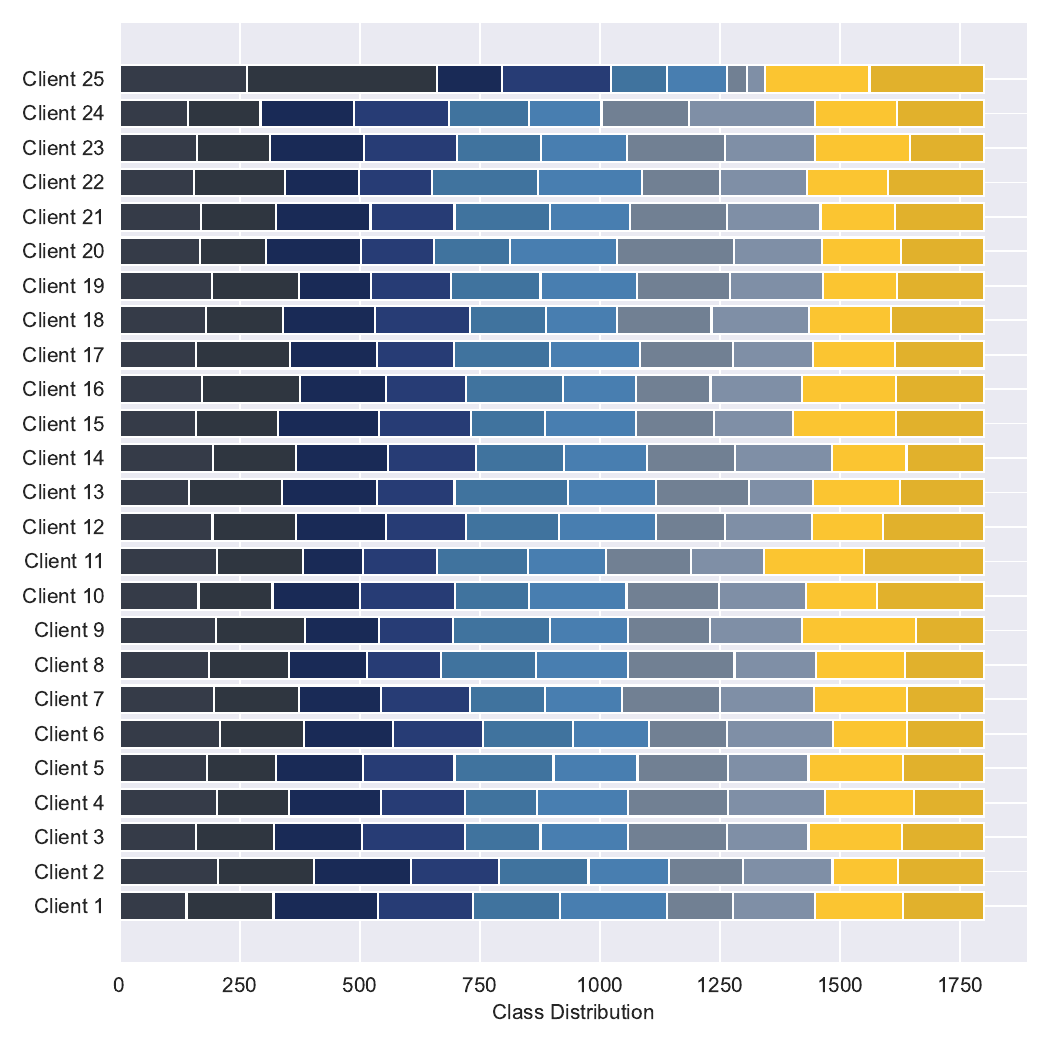}
        \caption{$\alpha = 100.0$}
        \label{app:fig:lda_100}
    \end{subfigure}%
    \hfill
    \begin{subfigure}{0.24\textwidth}
        \centering
        \includegraphics[width=\linewidth]{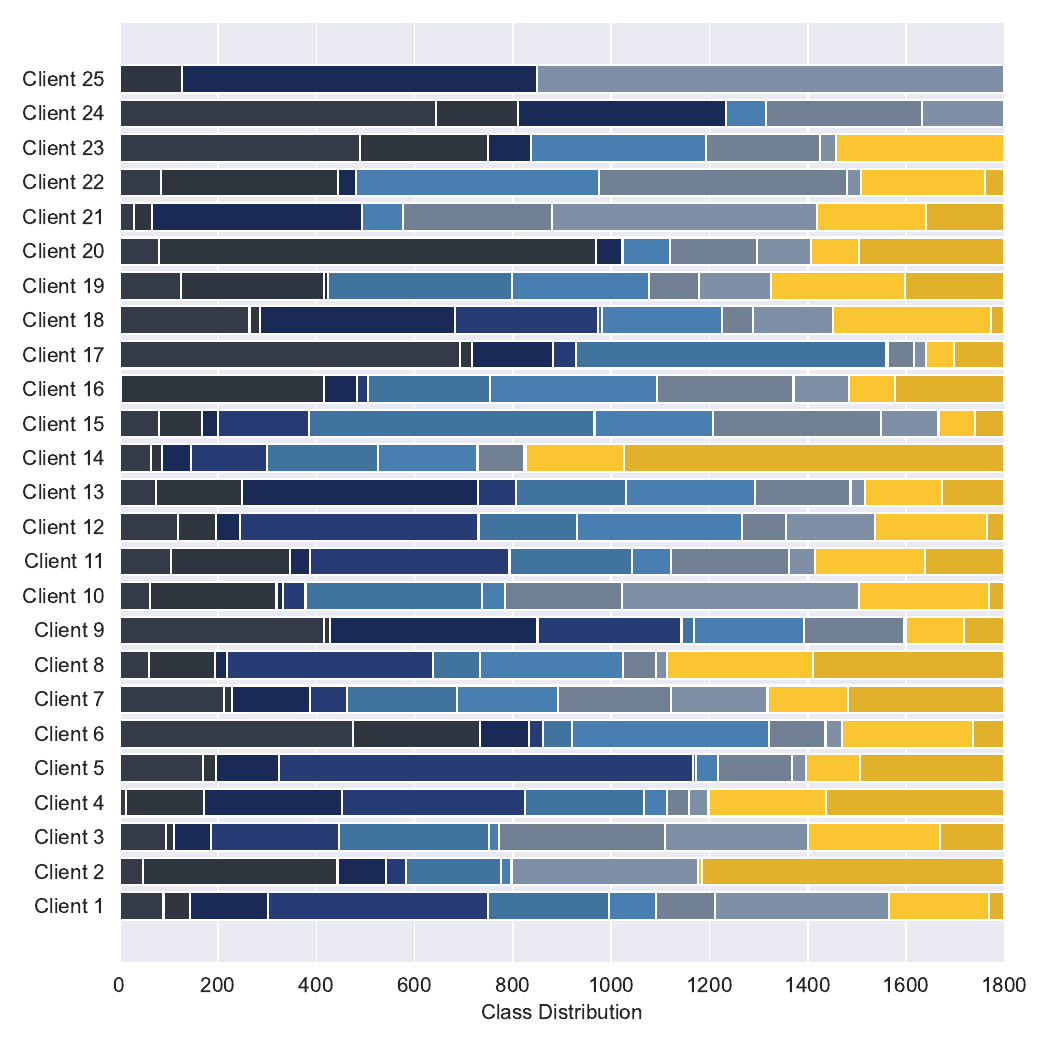}
        \caption{$\alpha = 1.0$}
        \label{app:fig:lda_1}
    \end{subfigure}%
    \hfill
    \begin{subfigure}{0.24\textwidth}
        \centering
        \includegraphics[width=\linewidth]{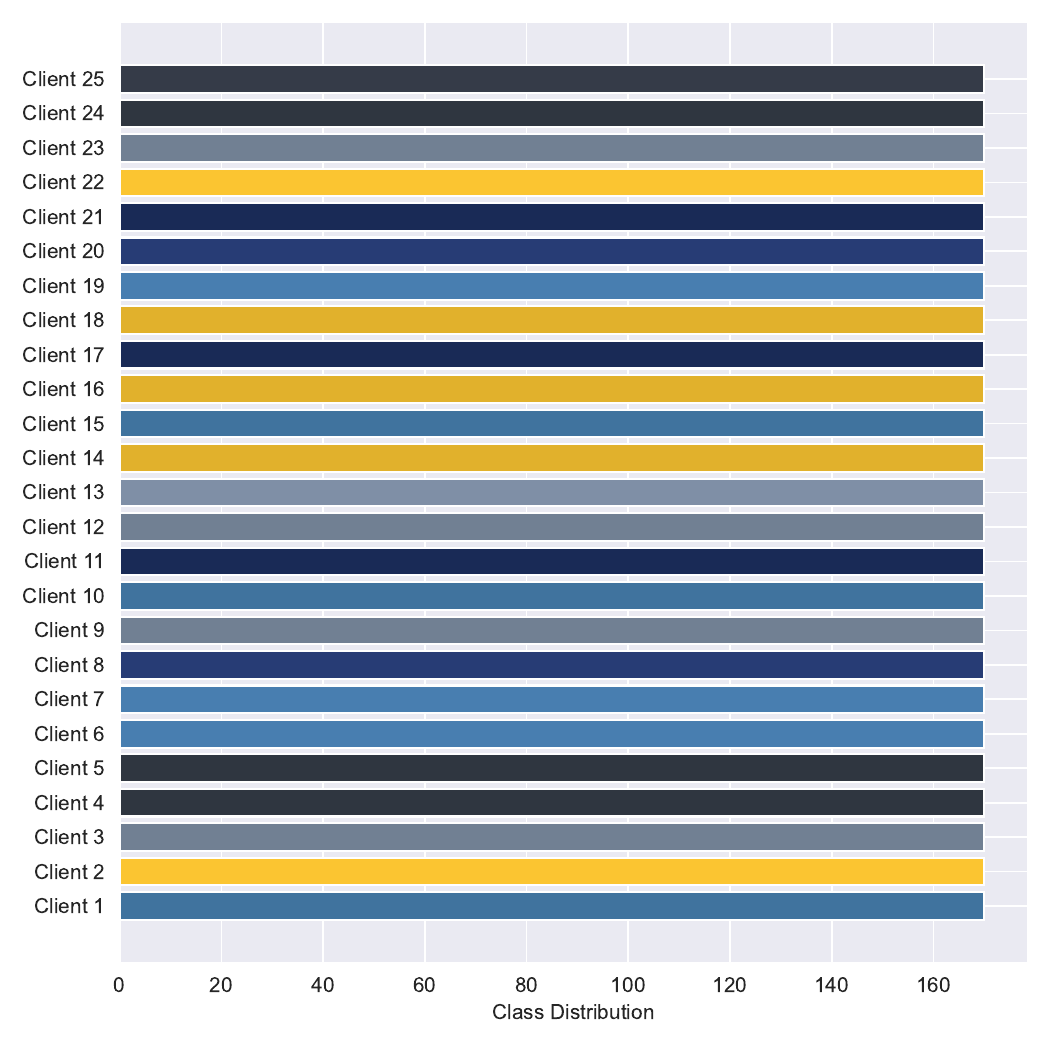}
        \caption{$\alpha \rightarrow 0.0$}
        \label{app:fig:lda_0}
    \end{subfigure}%
    \caption{Client data distribution induced by LDA Partitioning for differing $\alpha$ values.}
    \label{fig:lda_partitioning_different_alpha_values}
\end{figure}

The t-SNE for LDA partitioning is illustrated in \Cref{fig:tsne_lda_cifar100}.
As we decrease $\alpha$, the variable controlling client heterogeneity distinctly separates client embeddings.
This observation is aligned with \Cref{fig:lda_partitioning_different_alpha_values}, where more groups of clients share the same label distribution as $\alpha$ decreases.
In the IID scenario, all clients exhibit a uniform label distribution as $\alpha \rightarrow \infty$, resulting in a singular embedding cluster.

\begin{figure}[H]
    \centering
    \includegraphics[width=0.9\textwidth]{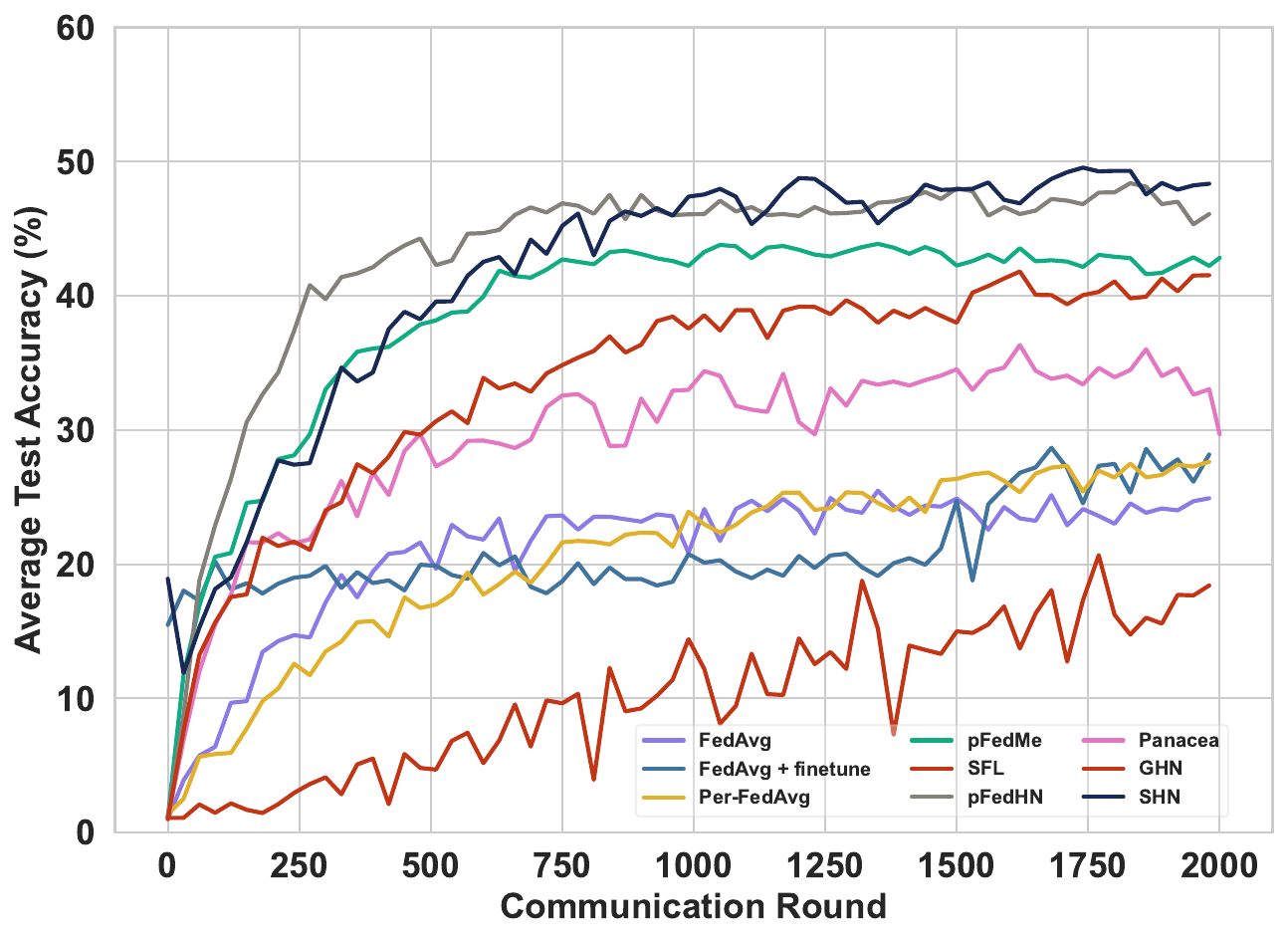}
    \caption{Average test accuracy for LDA partitioned CIFAR100 with $\alpha = 0.1$.}
    \label{app:fig:noniid_cifar100_results}
\end{figure}

\begin{figure}[H]
    \centering
    \includegraphics[width=0.9\textwidth]{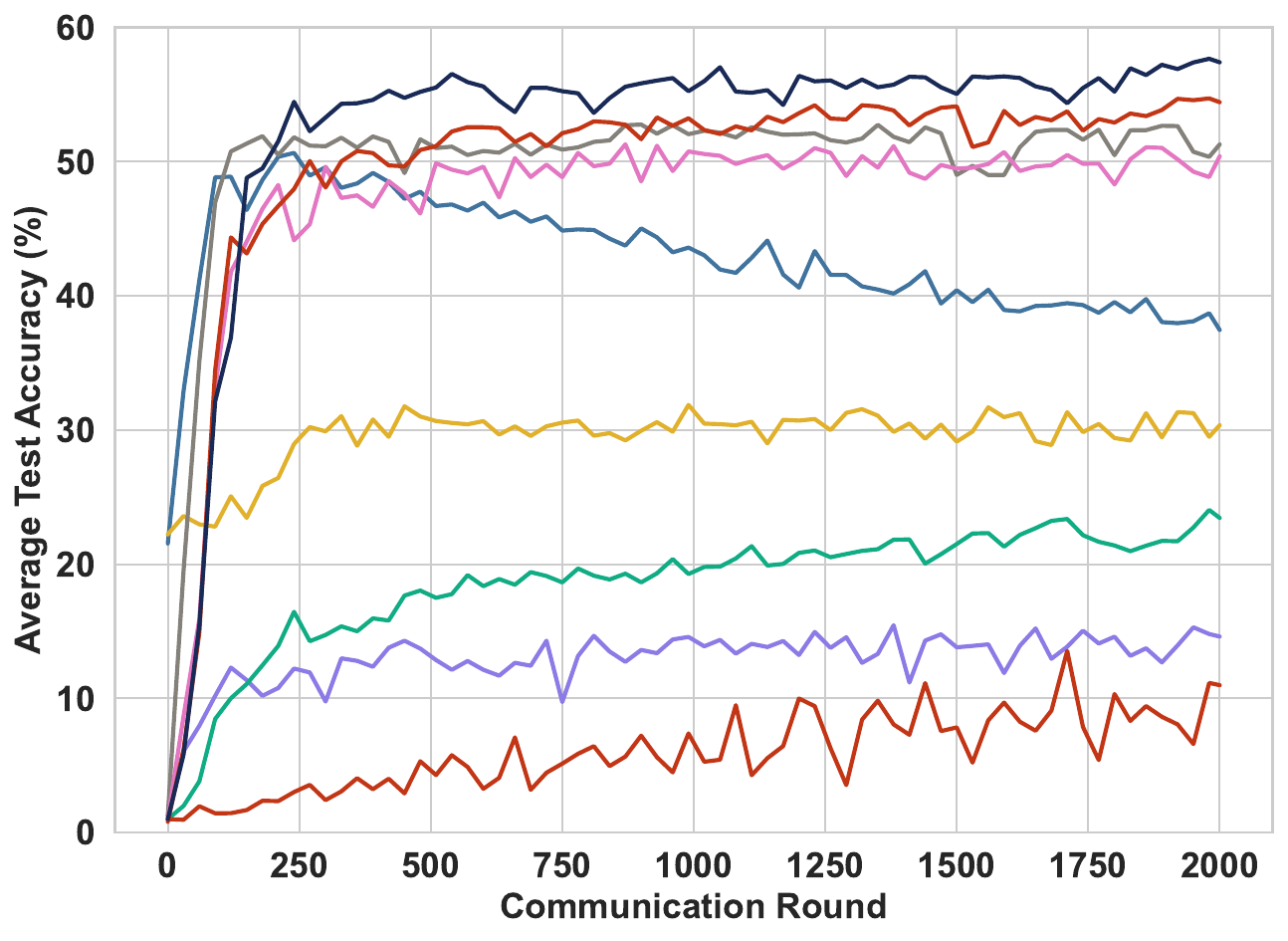}
    \caption{Average test accuracy for Cluster CIFAR100.}
    \label{app:fig:cluster_cifar100_results}
\end{figure}

\subsubsection{Federated Learning Results}
\label{appendix:federated_learning_results}

\textbf{Convergence.} \Cref{app:fig:noniid_cifar100_results,app:fig:cluster_cifar100_results} illustrate the convergence rate for all methods on non-iid CIFAR100 and Cluster CIFAR100.
In the non-iid CIFAR100 scenario, although SHN has a lower convergence rate than pFedHN, it begins to outperform pFedHN beyond $1000$ communication rounds.
This result is similar to Cluster CIFAR100, but SHN only requires $250$ communication rounds to outperform all baselines.
Learning the cellular sheaf causes the SHN to have a slower initial convergence rate.

\begin{figure}[H]
    \centering
    \begin{subfigure}{0.23\textwidth}
        \centering
        \includegraphics[width=\linewidth]{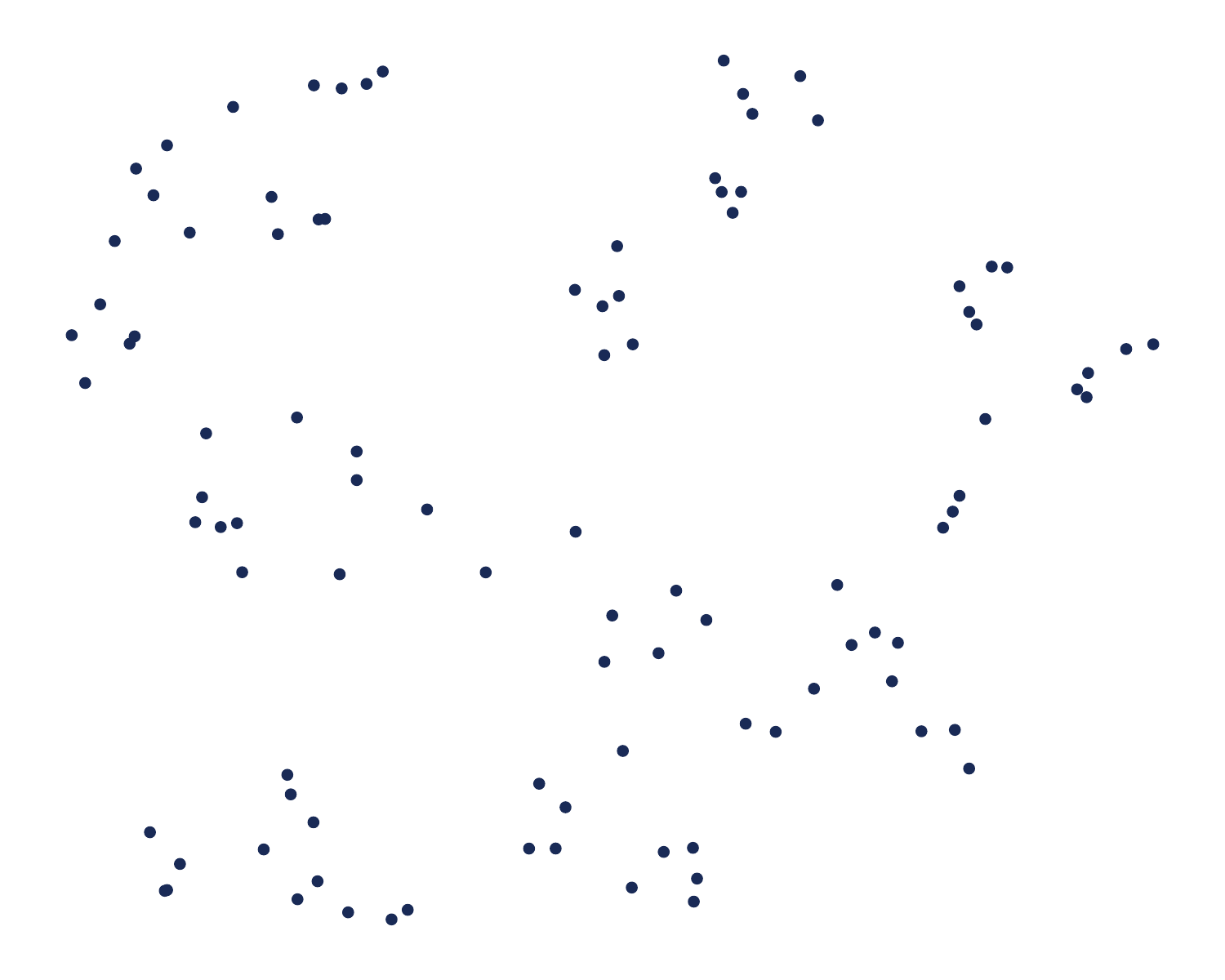}
        \caption{pFedHN}
    \end{subfigure}%
    \hfill
    \begin{subfigure}{0.23\textwidth}
        \centering
        \includegraphics[width=\linewidth]{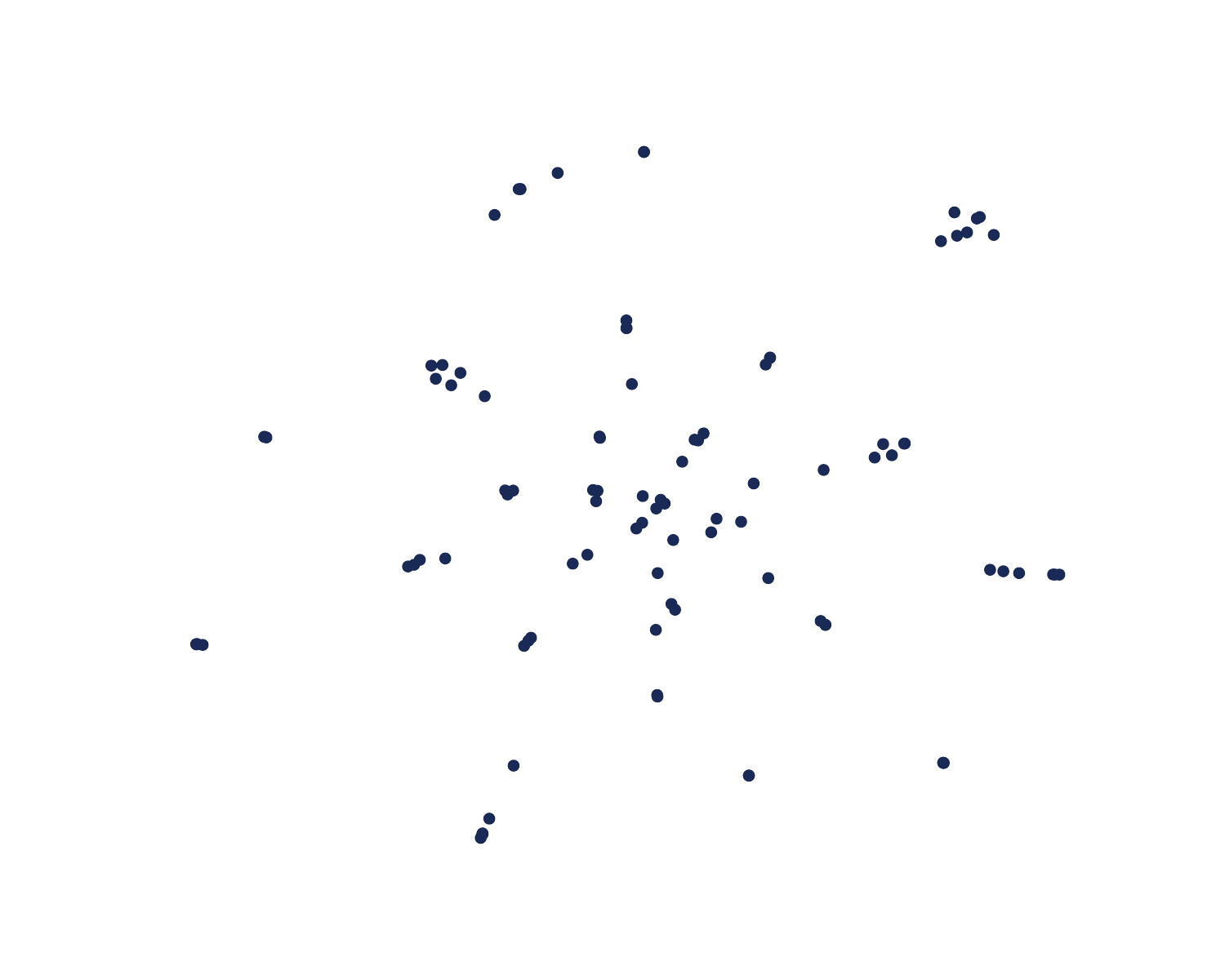}
        \caption{Panacea}
    \end{subfigure}%
    \hfill
    \begin{subfigure}{0.23\textwidth}
        \centering
        \includegraphics[width=\linewidth]{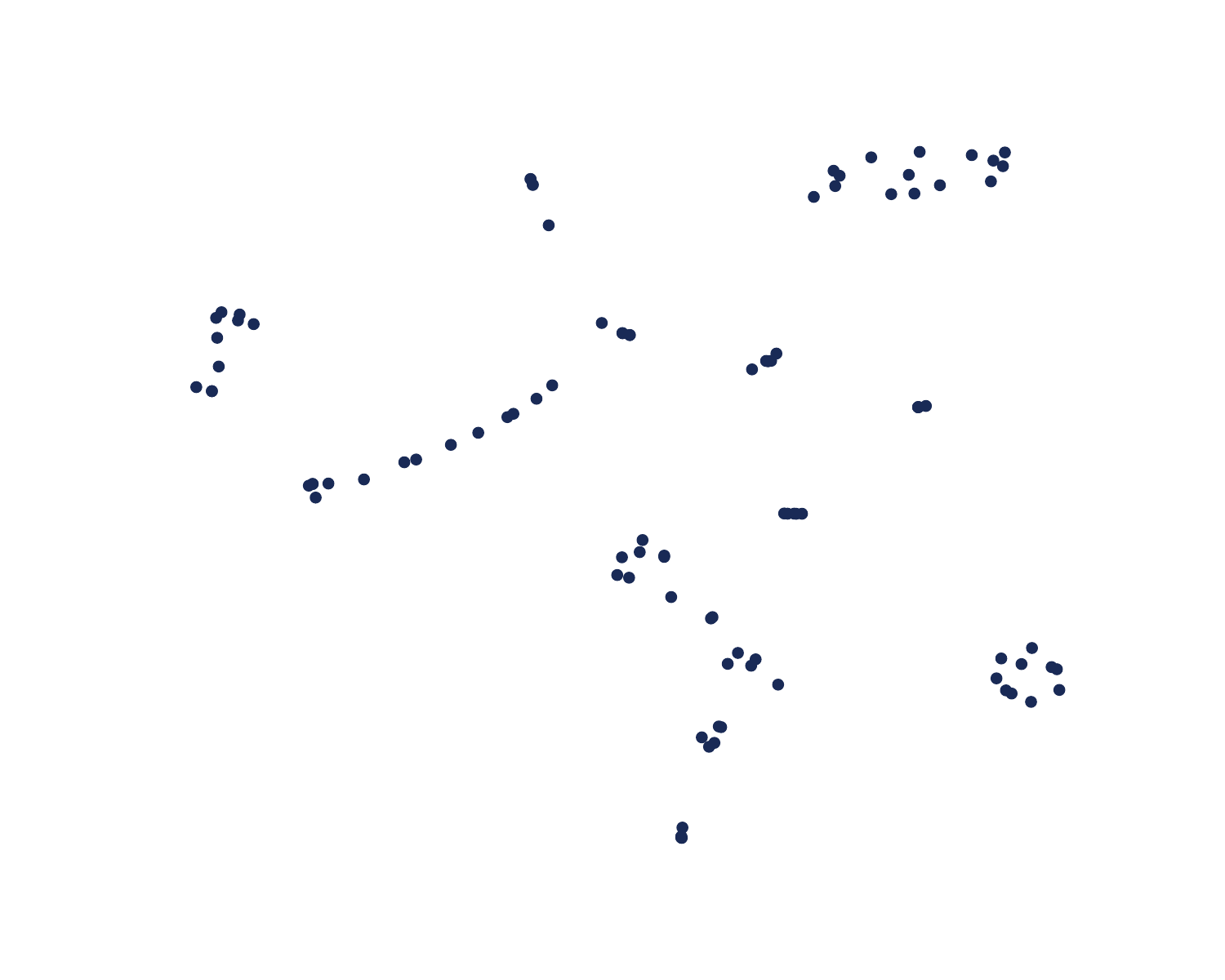}
        \caption{GHN}
    \end{subfigure}%
    \hfill
    \begin{subfigure}{0.23\textwidth}
        \centering
        \includegraphics[width=\linewidth]{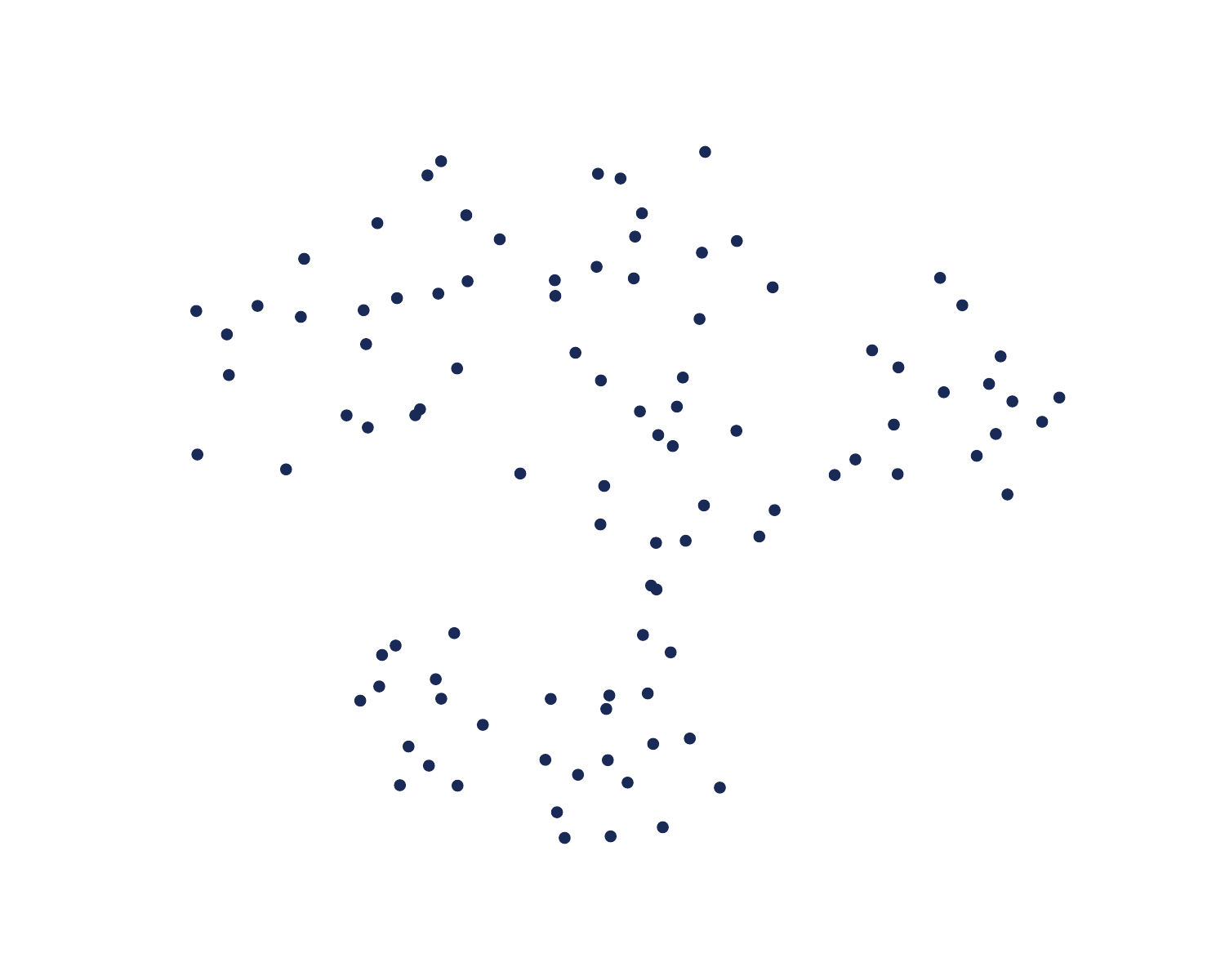}
        \caption{SHN}
    \end{subfigure}%
    \caption{t-SNE of client embeddings from cluster CIFAR100.}
    \label{fig:sh_cluster_cifar100_tsne}
\end{figure}

\begin{figure}[H]
    \centering
    \begin{subfigure}{0.23\textwidth}
        \centering
        \includegraphics[width=\linewidth]{images/results_and_discussion/01_lda_cifar100_hypernet.pdf}
        \caption{pFedHN}
    \end{subfigure}%
    \hfill
    \begin{subfigure}{0.23\textwidth}
        \centering
        \includegraphics[width=\linewidth]{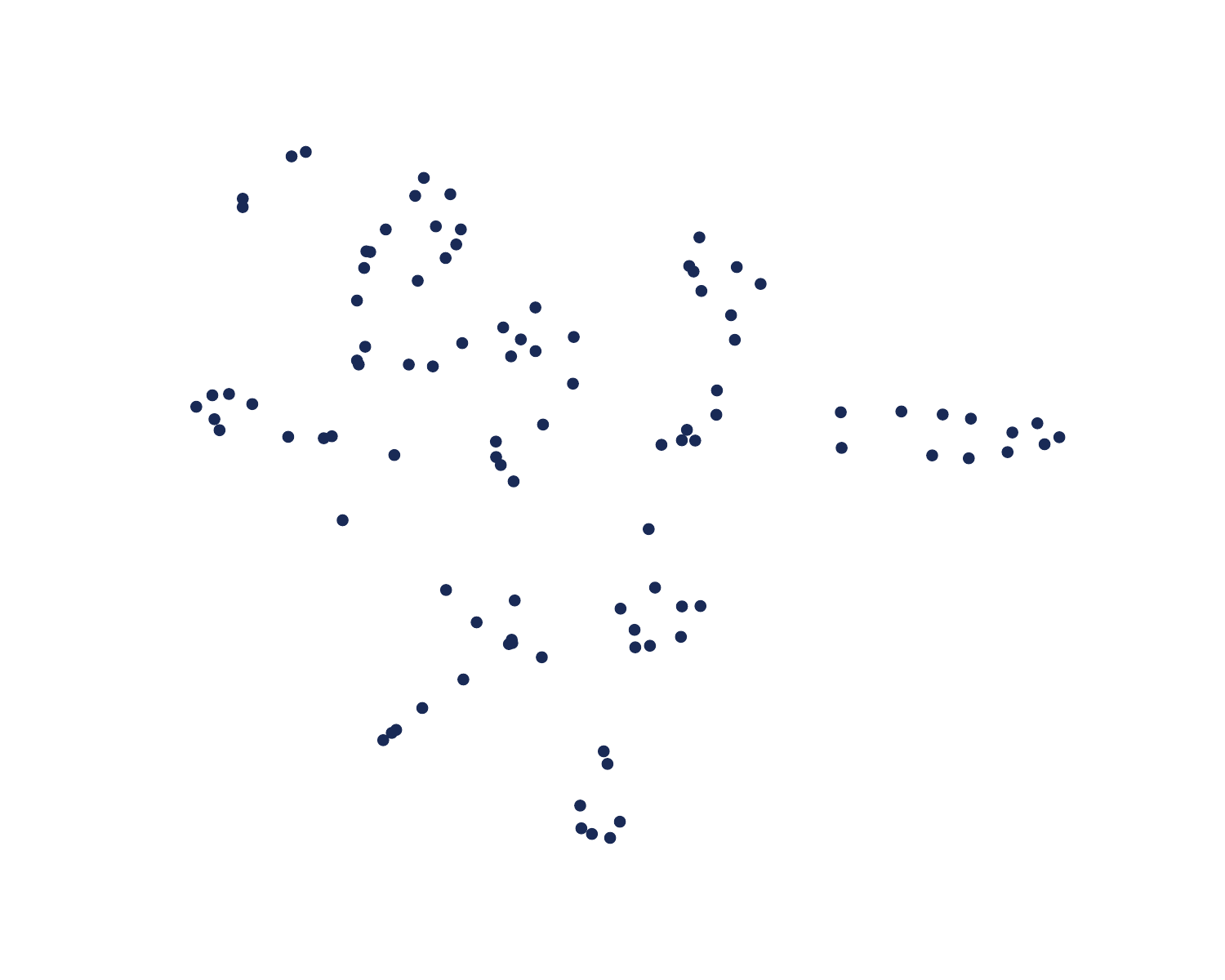}
        \caption{Panacea}
    \end{subfigure}%
    \hfill
    \begin{subfigure}{0.23\textwidth}
        \centering
        \includegraphics[width=\linewidth]{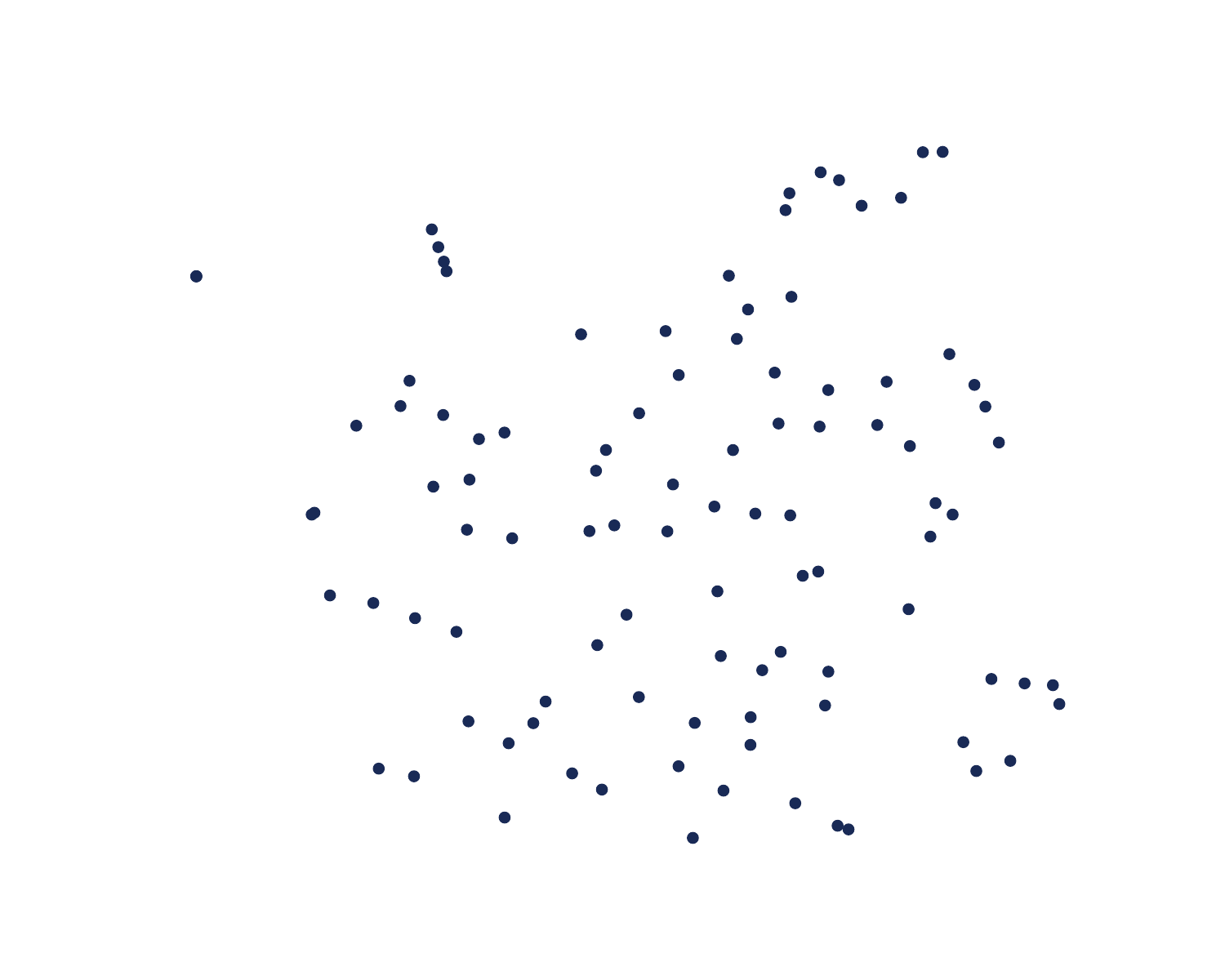}
        \caption{GHN}
    \end{subfigure}%
    \hfill
    \begin{subfigure}{0.23\textwidth}
        \centering
        \includegraphics[width=\linewidth]{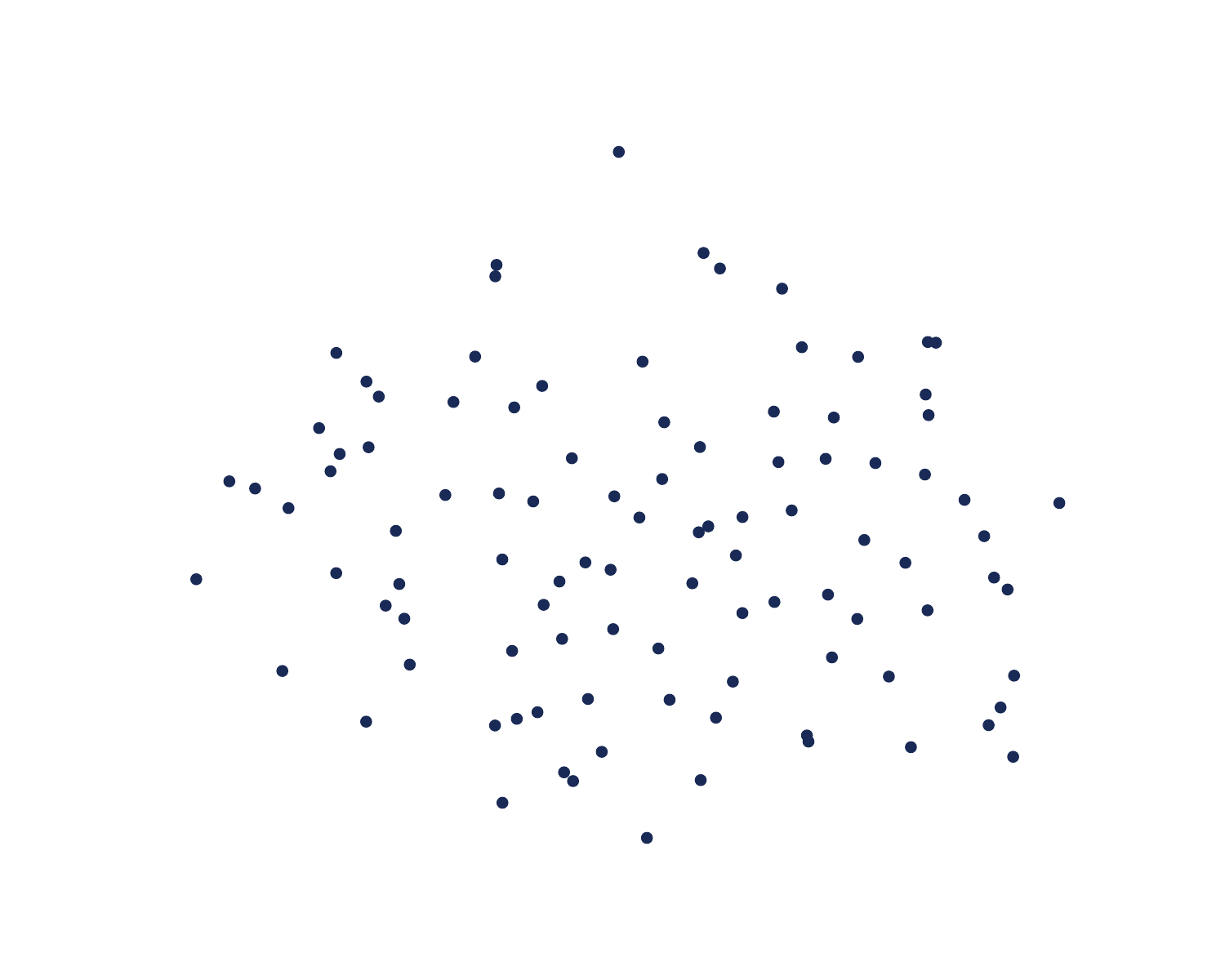}
        \caption{SHN}
    \end{subfigure}%
    \caption{t-SNE of client embeddings from non-IID CIFAR100 $\alpha = 0.1$}
    \label{fig:sh_non_iid_cifar100}
\end{figure}

\textbf{Learnt Client Embeddings.} To investigate the inner workings of our baselines, we explore the learned client embedding spaces from each model. Starting with pFedHN, the learned client embedding space is more continuous across clients. Conversely, Panacea, which incorporates a graph generator, imposes a more substantial relational inductive bias. This architectural distinction results in more clustering of embeddings, with groups displaying similar representations. GHN, operating under the same client relation graph as SHN, shapes its client embedding in alignment with the predefined graph structure. While SHN aligns with the client relation graph, it still learns a more continuous client embedding space, similar to pFedHN.

\subsubsection{Effectiveness of Sheaves}
\label{appendix:effectiveness_of_sheaves}

\begin{figure}[H]
    \centering
    \begin{subfigure}{0.335\textwidth}
        \centering
        \includegraphics[width=\linewidth]{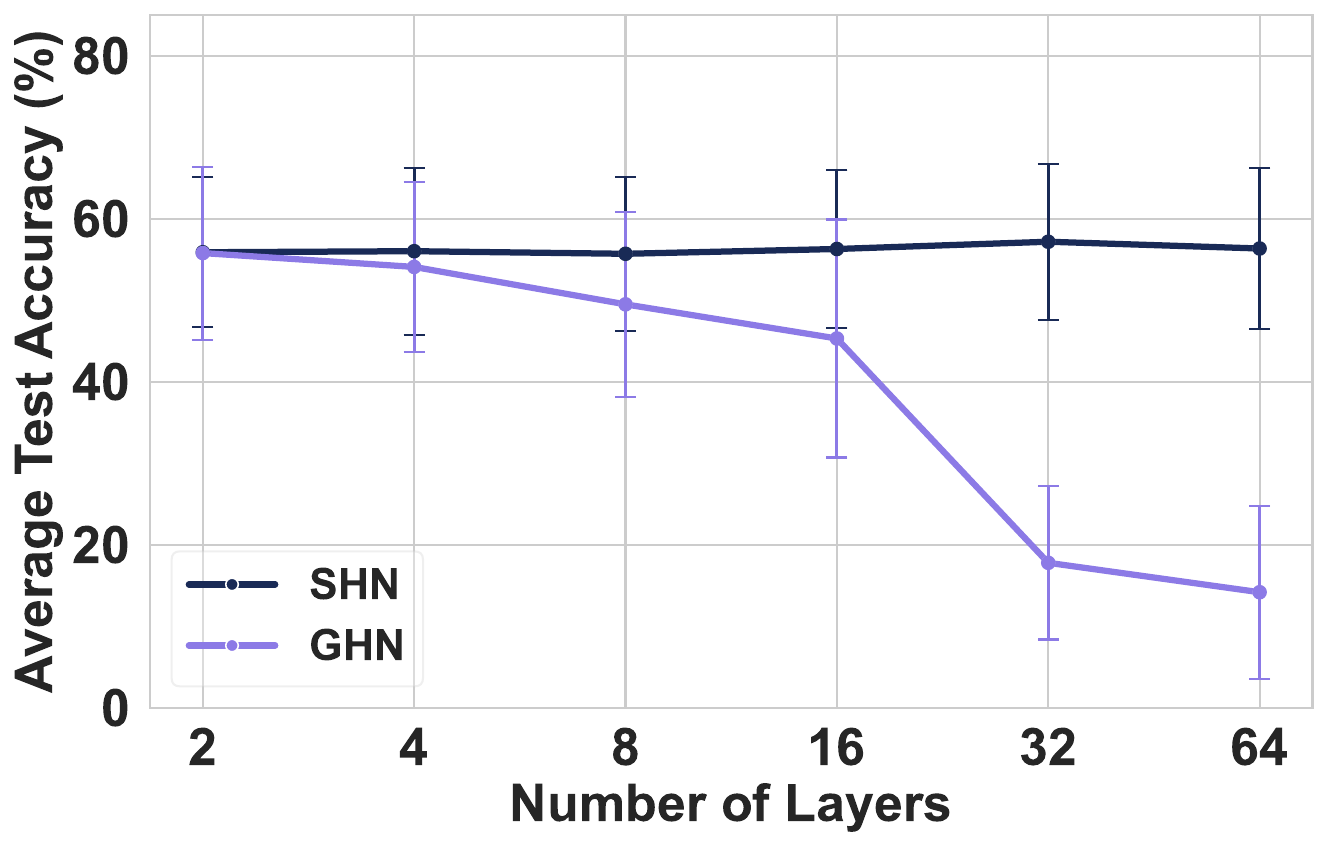}
        \caption{}
    \end{subfigure}
    \hfill
    \begin{subfigure}{0.32\textwidth}
        \centering
        \includegraphics[width=\linewidth]{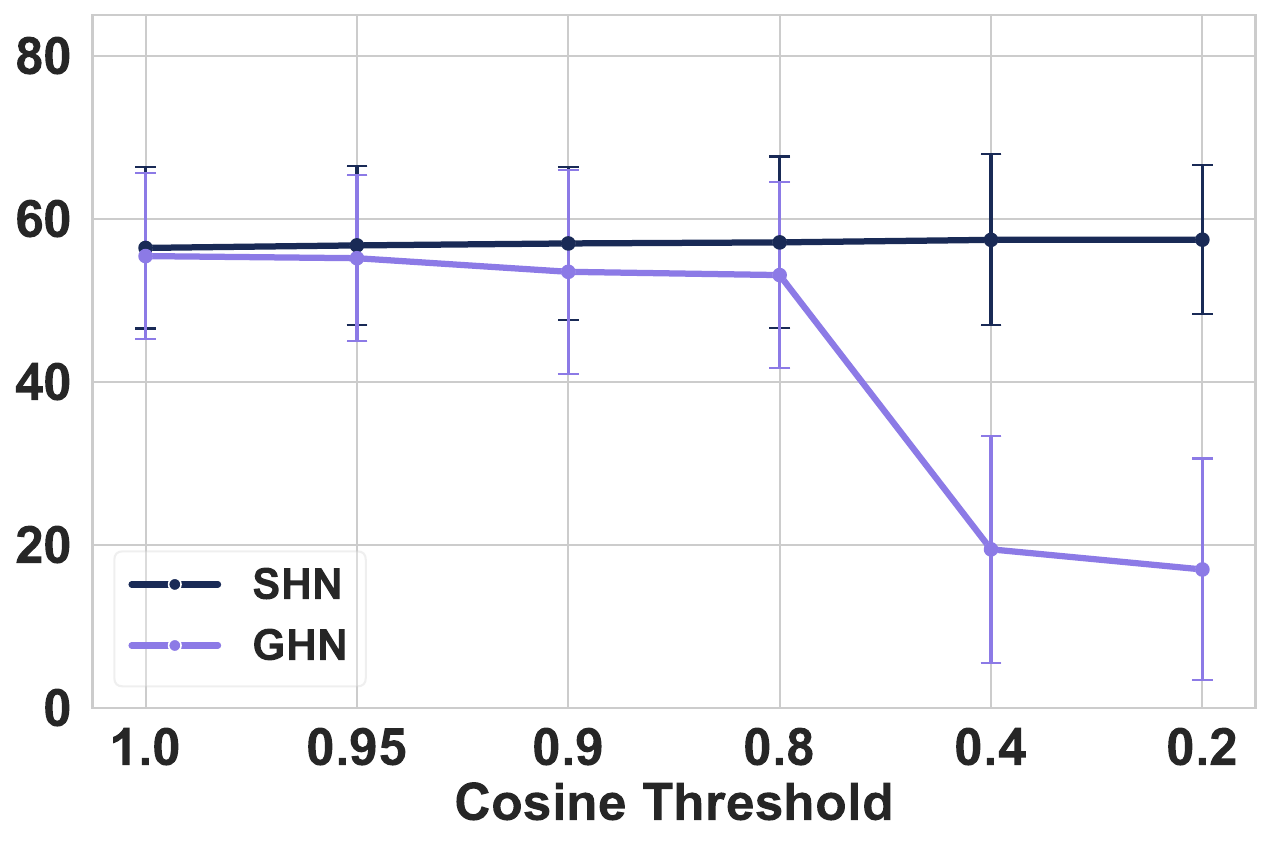}
        \caption{}
    \end{subfigure}
    \hfill
        \begin{subfigure}{0.32\textwidth}
        \centering
        \includegraphics[width=\linewidth]{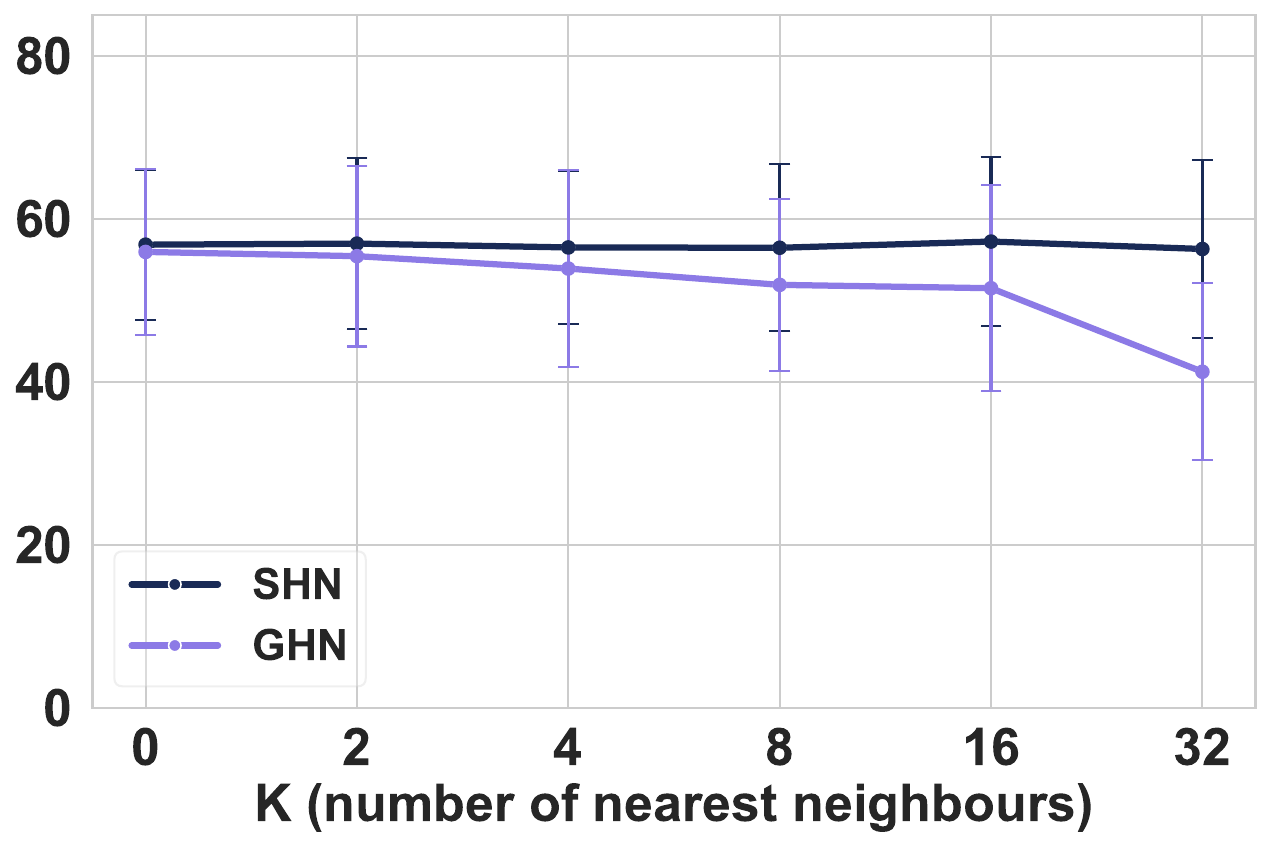}
        \caption{}
    \end{subfigure}
    \caption{\textbf{Personalized test accuracy \% on Cluster CIFAR100:} (d) As the number of message-passing layers increases. (e) As the cosine threshold decreases. (f) As the number of k-nearest neighbors increases.
    More details can be found in \Cref{appendix:effectiveness_of_sheaves}.
    }
    \label{app:fig:effectiveness_of_sheaves}
\end{figure}

\begin{table}[H]
\centering
\caption{Personalized client test accuracy and standard deviation on CIFAR100 ($\alpha = 0.1$) and Cluster CIFAR100 dataset as the number of message-passing layers increases. The client relation graph is constructed by $k$-nearest neighbors where $k = 3$.}
\vskip 0.11in
\begin{adjustbox}{max width=\textwidth}
\begin{tabular}{lccccccc}
\toprule
\textbf{Layers} & \textbf{2} & \textbf{4} & \textbf{8} & \textbf{16} & \textbf{32} & \textbf{64} & \textbf{Best}\\
\midrule
& \multicolumn{6}{c}{\textbf{CIFAR100 ($\alpha = 0.1$)}} \\
\cmidrule(lr){2-8}
GHN & 35.79 $\pm$ 12.07 & 32.22 $\pm$ 11.57 & 24.94 $\pm$ 10.15 & 19.96 $\pm$ 7.67 & 20.57 $\pm$ 7.31 & 18.29 $\pm$ 6.73 & 2\\
\textbf{SHN (ours)} & 44.94 $\pm$ 11.64 & 44.57 $\pm$ 11.13 & 46.14 $\pm$ 10.57 & 46.64 $\pm$ 10.92 & 46.41 $\pm$ 11.12 & 45.51 $\pm$ 10.03 & 16 \\
\midrule
& \multicolumn{6}{c}{\textbf{Cluster CIFAR100}} \\
\cmidrule(lr){2-8}
GHN & 55.82 $\pm$ 10.61 & 54.10 $\pm$ 10.43 & 49.51 $\pm$ 11.33 & 45.32 $\pm$ 14.60 & 17.81 $\pm$ 9.41 & 14.20 $\pm$ 10.59 & 2 \\
\textbf{SHN (ours)} & 55.91 $\pm$ 9.19 & 56.03 $\pm$ 10.23 & 55.71 $\pm$ 9.43 & 56.30 $\pm$ 9.65 & 57.19 $\pm$ 9.55 & 56.36 $\pm$ 9.82 & 32 \\
\bottomrule
\end{tabular}
\end{adjustbox}
\label{tab:num_layers_results}
\end{table}

\textbf{Number of Message-Passing Layers.} \Cref{tab:num_layers_results} outlines the performance of GHN and SHN as the number of message-passing layers exponentially increases. It shows that the additional expressivity of SHN, through its adoption of cellular sheaves, enables it to perform better than GHN when the number of layers is minimal and over-smoothing has not yet caused adverse effects. Additionally, as the number of layers increases, SHN maintains its performance, while GHN succumbs to over-smoothing, which worsens its performance.

\begin{table}[H]
\centering
\caption{Personalized test accuracy and standard deviation on the CIFAR100 ($\alpha = 0.1$) and Cluster CIFAR100 dataset as the number of $K$ nearest neighbors increases. The number of message-passing layers for both GHN and SHN is fixed at $3$.}
\vskip 0.11in
\begin{adjustbox}{max width=\textwidth}
\begin{tabular}{lccccccc}
\toprule
\bf$K$ & \textbf{0} & \textbf{2} & \textbf{4} & \textbf{8} & \textbf{16} & \textbf{32} & \textbf{Best}\\
\midrule
& \multicolumn{6}{c}{\textbf{CIFAR100 ($\alpha = 0.1$)}} \\
\cmidrule(lr){2-8}
GHN & 41.71 $\pm$ 12.04 & 34.36 $\pm$ 12.24 & 32.93 $\pm$ 11.41 & 30.29 $\pm$ 9.44 & 27.05 $\pm$ 9.58 & 20.05 $\pm$ 8.71 & 0 \\
\textbf{SHN (ours)} & 45.74 $\pm$ 11.34 & 46.14 $\pm$ 11.34 & 46.74 $\pm$ 10.41 & 46.41 $\pm$ 11.26 & 47.52 $\pm$ 10.36 & 47.47 $\pm$ 10.42 & 16 \\
\midrule
& \multicolumn{6}{c}{\textbf{Cluster CIFAR100}} \\
\cmidrule(lr){2-8}
GHN & 55.74 $\pm$ 10.14 & 55.42 $\pm$ 11.08 & 53.9 $\pm$ 12.11 & 51.9 $\pm$ 10.54 & 51.49 $\pm$ 12.61 & 41.24 $\pm$ 10.86 & 0 \\
\textbf{SHN (ours)} & 56.83 $\pm$ 9.22 & 56.96 $\pm$ 10.52 & 56.49 $\pm$ 9.41 & 56.45 $\pm$ 10.22 & 57.21 $\pm$ 10.38 & 56.30 $\pm$ 10.87 & 16 \\
\bottomrule
\end{tabular}
\end{adjustbox}
\label{tab:knn_results}
\end{table}

\begin{table}[H]
\centering
\caption{Personalized test accuracy and standard deviation for the CIFAR100 ($\alpha = 0.1$) and Cluster CIFAR100 datasets, as the cosine threshold value declines. The number of message-passing layers for both GHN and SHN is fixed at $3$.}
\vskip 0.11in
\begin{adjustbox}{max width=\textwidth}
\begin{tabular}{lccccccc}
\toprule
\textbf{Cosine Threshold} & \textbf{1.0} & \textbf{0.95} & \textbf{0.9} & \textbf{0.8} & \textbf{0.4} & \textbf{0.2} & \textbf{Best}\\
\midrule
& \multicolumn{6}{c}{\textbf{CIFAR100 ($\alpha = 0.1$)}} \\
\cmidrule(lr){2-8}
GHN & 40.91 $\pm$ 11.08 & 43.07 $\pm$ 11.19 & 41.84 $\pm$ 11.49 & 43.35 $\pm$ 11.91 & 36.34 $\pm$ 12.01 & 27.37 $\pm$ 10.56 & 0.8 \\
\textbf{SHN (ours)} & 42.74 $\pm$ 12.02 & 44.45 $\pm$ 10.74 & 46.28 $\pm$ 11.03 & 45.51 $\pm$ 11.01 & 45.12 $\pm$ 11.72 & 45.24 $\pm$ 10.53 & 0.9 \\
\midrule
& \multicolumn{6}{c}{\textbf{Cluster CIFAR100}} \\
\cmidrule(lr){2-8}
GHN & 55.43 $\pm$ 10.21 & 55.17 $\pm$ 10.18 & 53.50 $\pm$ 12.51 & 53.10 $\pm$ 11.42 & 19.46 $\pm$ 13.89 & 16.99 $\pm$ 13.61 & 1\\
\textbf{SHN (ours)} & 56.41 $\pm$ 9.88 & 56.74 $\pm$ 9.71 & 56.98 $\pm$ 9.39 & 57.11 $\pm$ 10.53 & 57.42 $\pm$ 10.49 & 57.44 $\pm$ 9.16 & 0.2 \\
\bottomrule
\end{tabular}
\end{adjustbox}
\label{tab:cosine_threshold_results}
\end{table}

\textbf{Number of Nearest Neighbors and Cosine Threshold.} \Cref{tab:knn_results} and \Cref{tab:cosine_threshold_results} present the performance of GHN and SHN under different settings of the client relation graph, with the number of message-passing layers fixed at $3$ for both models. As the number of nearest neighbors increases or the cosine threshold declines, more connections between clients are established, resulting in a more heterophilic setting.
Both tables show that GHN's performance declines under these conditions while SHN remains robust.
Otherwise, the results demonstrate that SHN is insensitive to these hyperparameters and operates effectively on sub-optimal graphs.

\subsection{Experimental Setup}
\label{appendix:experimental_setup}

In this section, we provide a detailed overview of our experimental setup. In \Cref{appendix:datasets}, we describe the PFL challenges each dataset is designed to probe and explain how they have been partitioned across the client pool. \Cref{appendix:latent_dirichlet_allocation} illustrates the impact of varying values of $\alpha$ on the local client data distribution. The number of parameters for each client model used in our experiments is outlined in \Cref{appendix:client_model_parameters}. The personalized evaluation metric used to present our results is discussed in \Cref{appendix:evaluation_metric}. Finally, \Cref{appendix:training_details} presents the hyperparameter search space used to train the SHN.

\subsubsection{Datasets}
\label{appendix:datasets}

\begin{table}[H]
\centering
\caption{\textbf{Summary of datasets.}}
\label{table:summary_of_datasets}
\begin{adjustbox}{max width=\textwidth}
\begin{tabular}{llllll}
\toprule
\textbf{Dataset}          & \textbf{Domain}                     & \textbf{Num Clients/Nodes} & \textbf{Num Edges} & \textbf{Client Model Type} & \textbf{Client Local Objective} \\
\midrule
CIFAR100 \cite{krizhevsky_learning_nodate}        & Multi-class Classification & 100                & Variable* & CNN               & Classification         \\
Cluster CIFAR100 \cite{shamsian_personalized_2021} & Multi-class Classification & 100                & Variable* & CNN               & Classification         \\
CompCars \cite{noauthor_compcars_nodate}        & Vehicle Classification     & 30                 & 158                           & ResNet-18               & Classification         \\
METR-LA \cite{metr_la_dataset}         & Traffic Forecasting        & 207                & 1722                          & GRU               & Regression             \\
PEMS-BAY \cite{pems_bay_dataset}        & Traffic Forecasting        & 325                & 2694                          & GRU               & Regression             \\
TPT-48 \cite{post_washingtonpostdata-2c-beyond--limit-usa_2024}          & Weather Forecasting        & 48                 & 258                           & GRU               & Regression             \\
\bottomrule
\end{tabular}
\label{t}
\end{adjustbox}
\end{table}

\textbf{CIFAR100}. Our evaluation includes a classic Federated Learning (FL) benchmark dataset, CIFAR100. To partition the data, we employ Latent Dirichlet Allocation (LDA) to partition the data, as further detailed in \Cref{appendix:latent_dirichlet_allocation}, using two distinct $\alpha$ values: $1000$ for simulating an IID setting and $0.1$ for a non-IID scenario. Although LDA partitioned CIFAR100 does not inherently possess a strong relational topology among clients, they allow us to evaluate the same standard adopted by other methods in the literature. 

\textbf{Cluster CIFAR100.} To compose a dataset with a strong topological structure, we create the Cluster CIFAR100 dataset, where each client is assigned to one of the $20$ superclasses. This dataset is inherently heterogeneous and evaluates not just the capability of SHNs to deliver personalized parameters but also their ability to mitigate detrimental effects from inappropriate parameter aggregation. For instance, conflicting signals from disparate features learned by models in different classes can compromise model performance when aggregated. Effective PFL methods should either circumvent such detrimental aggregation or employ advanced aggregation techniques that preserve the integrity of personalized learning.

The following datasets were constructed from the same code path, provided by \citet{lin_graph-relational_2023}, to ensure consistency when comparing across all baselines. The FL challenges each dataset is designed to probe and how they are partitioned are as follows.

\textbf{CompCars.} Advancing to a more complex and large-scale computer vision challenge, we incorporate the Comprehensive Cars dataset into our evaluation framework. In this dataset, clients possess images of vehicles from identical manufacturing years and viewpoints. This setup allows us to examine how PFL methods scale with the increase in client model parameters, offering insights into the scalability of SHNs.

This dataset contains $136,726$ images of cars with labels describing their type, year of manufacture (YOM), and viewpoint. The dataset was split with the same methodology from \citet{xu_graph-relational_2023}. A subset of the dataset was used, where four car types (MPV, SUV, sedan, and hatchback) were selected. Following this, the dataset is partitioned so that each client holds images of cars manufactured in the same year and viewpoint. A client graph is constructed by connecting clients with nearby viewpoints or YOMs. For example, two clients are connected if their viewpoints are \say{front} and \say{front-side}, respectively.

\textbf{PEMS-BAY \& METR-LA.} Our evaluation includes two traffic forecasting datasets, PEMS-BAY \cite{pems_bay_dataset} and METR-LA \cite{metr_la_dataset}, where traffic sensors, serving as clients, are tasked with predicting traffic flow for the next 60 minutes. These datasets consist of readings from individual traffic sensors monitoring vehicle speeds at various intervals. To adapt these datasets to a federated learning framework, they are modelled as graphs where sensors are represented as nodes (clients) and are connected based on their physical proximity. Such a graph structure allows us to leverage client relationships to enhance forecasting accuracy.

PEMS-BAY consists of $325$ traffic sensors in the Bay Area (US) with $6$ months of traffic readings between the $1$st of January and $31$st of May 2017 in $5$ minute intervals. METR-LA contains traffic readings from $207$ sensors in Los Angeles County over a $4$ month period between $1$st March 2012 to $30$th June 2012. For both datasets, we construct a client relation graph from the same methodology provided by \citet{meng_cross-node_2021}, where each sensor is a node and two sensors are connected based on their distance.

\textbf{TPT-48.}  In the context of weather forecasting, we assess the performance of our methods using the \emph{TPT-48} dataset, which involves predicting the temperature in 48 US states over the next six months, where each state acts as a client. In the context of a graph, each state is represented as a node, and edges connect geographically adjacent states. Despite their proximity, the climatic conditions across these states vary considerably, leading to a heterophilic graph structure. This benchmark is crucial for evaluating the capability of PFL methods to handle heterophily effectively, potentially improving performance by utilizing meaningful but complex inter-state relationships. We use the data processed by Washington Post \cite{post_washingtonpostdata-2c-beyond--limit-usa_2024}.

\subsubsection{Latent Dirichlet Allocation}
\label{appendix:latent_dirichlet_allocation}

To simulate a realistic cross-device FL environment within the CIFAR100 dataset, we induce heterogeneity in our clients by employing Latent Dirichlet Allocation (LDA). LDA is a generative probabilistic model widely used for analyzing collections of discrete data. By manipulating the concentration hyper-parameter $\alpha$, we can control the degree of heterogeneity projected by our clients in the system. In the case of $\alpha \rightarrow 0.0$, we obtain a highly heterogeneous distribution across our clients, where clients exhibit distinct label spaces. Conversely, as $\alpha \rightarrow \inf$, the system tends towards homogeneity, resulting in a uniform label space across all clients.

\subsubsection{Client Model Parameters}
\label{appendix:client_model_parameters}

Our experiments involved $4$ different client model architectures of various sizes. \Cref{table:client_model_parameters} describes each client model architecture, the task they were used for, and the number of learnable parameters. While our experiments were not conducted on models consisting of billions of parameters, we hope to illustrate that our evaluation was performed on various model architectures.

\begin{table}[H]
\caption{\textbf{Number of Parameters for each client model.}}
\centering
\begin{tabular}{llc}
\toprule
\textbf{Model} & \textbf{Task} & \textbf{Number of Parameters} \\ \midrule
CNN & Classification & 128,832 \& 90,242\\
ResNet-18 & Classification & 11,178,564\\
MLP & Regression & 354\\
GRU & Regression & 26,177\\
\bottomrule
\end{tabular}
\label{table:client_model_parameters}
\end{table}

\subsubsection{Evaluation Metric}
\label{appendix:evaluation_metric}

We present the results of our PFL experiments as the \textbf{personalized} test accuracy or mean squared error. Each client evaluates their local test set and reports the performance of the personalized parameters to the server. Subsequently, the mean and standard deviation were measured across \textbf{all} clients. Each experiment was independently performed $5$ times, and the average means and standard deviations across all experiments are reported. \Cref{alg:personalized_evaluation} details our evaluation strategy for a single experiment.

\begin{algorithm}[H]
    \DontPrintSemicolon
    \KwIn{SHN parameters - $\psi_{\mathcal{H}}$, \emph{stalk} dimension - $d$, Client relation graph, adjacency matrix, and node embeddings: G, A, X}
    \KwOut{Average $\mu$ and $\sigma$ personalized test performance}
    \BlankLine
    $\mu \gets 0.0$\;
    $\sigma \gets 0.0$\;
    \For{\textup{each client} $m = 1, 2, \ldots, |M|$}{
        $\theta_{m} \gets \mathcal{SHN}(m, d, G, A, X; \psi_{\mathcal{H}})$\;
        $\mu_m, \sigma_m \gets $ EvaluateClient($m$, $\theta_{m}$)\;
        $\mu \gets \mu + \mu_m$\;
        $\sigma \gets \sigma + \sigma_m$\;
    }

    $\mu \gets \mu / |M|$\;
    $\sigma \gets \sigma / |M|$\;
    
    \Return{$\mu, \sigma$}
    \caption{Personalized Evaluation}
    \label{alg:personalized_evaluation}
\end{algorithm}

\subsubsection{Training Details}
\label{appendix:training_details}

The hyperparameter search space for training is provided in \Cref{tab:sheaf_hypernetwork_hyperparameter_search_space}. The training was performed on $4$ NVIDIA GeForce RTX $2080$s and $2$ NVIDIA A40s, each device having $8$ and $48$ GB of GPU memory, respectively. 

\begin{table}[H]
\centering
\caption{Sheaf HyperNetwork hyper-parameter search space.}
\begin{tabular}{ll}
\toprule
\textbf{Hyper-parameter} & \textbf{Values}                 \\ \hline
Hidden channels          & \{8, 16, 32, 64\}                 \\
Stalk dimension $d$      & \{1, 2, 3, .. 5\}                 \\
Layers                   & \{1, 2, 4, 8, 16, 32, 64\}        \\
Activation               & Tanh, ELU                       \\
Optimizer                & Adam                            \\
Learning Rate            & 0.001                           \\
Weight Decay             & 0.00005                         \\
Cosine Similarity        & \{1.0, 0.95, 0.9, 0.8, 0.4, 0.2\} \\
K-Nearest Neighbours     & \{0, 2, 4, 8, 16, 32\}            \\
\bottomrule
\end{tabular}
\label{tab:sheaf_hypernetwork_hyperparameter_search_space}
\end{table}

\end{document}